\documentclass{article}

\usepackage{microtype}
\usepackage{graphicx}
\usepackage{booktabs}
\usepackage{amsmath,amssymb}
\usepackage{float}

\usepackage{amsmath,amsfonts,bm}

\def\eqref#1{equation~\ref{#1}}
\def\1{\bm{1}}

\DeclareMathAlphabet{\mathsfit}{\encodingdefault}{\sfdefault}{m}{sl}
\SetMathAlphabet{\mathsfit}{bold}{\encodingdefault}{\sfdefault}{bx}{n}

\newcommand{\TeacherRetrievalRange}{0.89 to 1.00}
\newcommand{\TeacherAggRange}{0.87 to 0.93}
\newcommand{\TeacherMultihopRange}{0.95 to 1.00}
\newcommand{\TeacherMultihopFailed}{0.07}
\newcommand{\RecoveryStreamWrongPct}{1.6\%}
\newcommand{\CostSixteenMinutes}{17}

\newcommand{\CostTwelveEightMinutes}{150}
\newcommand{\CostTwelveEightExamples}{600}
\newcommand{\CostMillionDays}{6}
\newcommand{\CostMillionDaysTwelveEight}{174}
\newcommand{\ExpertBothLo}{0.39}
\newcommand{\ExpertBothHi}{0.71}
\newcommand{\ExpertRoutedLo}{0.72}
\newcommand{\ExpertRoutedHi}{0.87}
\newcommand{\BPGoldKlLatest}{$+0.43$ [$+0.35$, $+0.50$]}
\newcommand{\BPGoldKlTwohop}{$+0.34$ [$+0.27$, $+0.42$]}
\newcommand{\BPRhoKlLo}{$0.50$}
\newcommand{\BPRhoKlHi}{$0.67$}
\newcommand{\BPRhoGoldLo}{$-0.29$}
\newcommand{\BPRhoGoldHi}{$0.14$}
\newcommand{\BPTimingMax}{0.32}
\newcommand{\UniqueFourGapPts}{11}

\newcommand{\MaskedSetRecLo}{-0.12}
\newcommand{\MaskedSetRecHi}{-0.08}

\newcommand{\GranDirectNativeTwoFive}{$-0.37$ [$-0.44$, $-0.29$]}
\newcommand{\GranSetRecTwoFive}{$-0.07$ [$-0.15$, $+0.01$]}
\newcommand{\GranDirectTwoFive}{0.49}

\newcommand{\SentAnnTwoFive}{$+0.06$ [$-0.02$, $+0.13$]}
\newcommand{\SentAnnFifty}{$+0.08$ [$+0.02$, $+0.14$]}
\newcommand{\SentKzTwoFive}{$+0.07$ [$+0.00$, $+0.13$]}
\newcommand{\SentKzFifty}{$+0.07$ [$+0.02$, $+0.12$]}
\newcommand{\SentN}{234}
\newcommand{\SentKeptPct}{2}
\newcommand{\SentPilotOutside}{45}
\newcommand{\KzStaleCausalLo}{0.95}
\newcommand{\KzStaleCausalHi}{1.00}
\newcommand{\KzStaleHtwoLo}{0.20}
\newcommand{\KzStaleHtwoHi}{0.57}

\newcommand{\AllYesNoN}{89}
\newcommand{\AllExtractiveN}{569}
\newcommand{\AllRaFOneTwoFive}{$+0.06$ [$+0.04$, $+0.08$]}
\newcommand{\AllRaContainTwoFive}{$+0.08$ [$+0.04$, $+0.11$]}
\newcommand{\AllRaYesNoTwoFive}{$-0.01$ [$-0.10$, $+0.08$]}
\newcommand{\AllRaFOneFifty}{$+0.03$ [$+0.02$, $+0.05$]}
\newcommand{\AllRkFOneTwoFive}{$-0.10$ [$-0.11$, $-0.08$]}
\newcommand{\AllRkFOneFifty}{$-0.05$ [$-0.07$, $-0.04$]}
\newcommand{\QasperRows}{658}
\newcommand{\QasperDocs}{261}
\newcommand{\QasperDense}{234}
\newcommand{\QasperCondDocs}{161}
\newcommand{\QasperTrainQs}{1443}
\newcommand{\QasperTrainDocs}{693}
\newcommand{\QasperTrainSolved}{541}
\newcommand{\QasperTrainLabeledQs}{401}
\newcommand{\QasperTrainLabeledDocs}{306}
\newcommand{\KvzipCohortN}{238}
\newcommand{\KvzipCohortDocs}{164}
\newcommand{\KvzipTwoFive}{0.86}
\newcommand{\KvzipFifty}{0.92}
\newcommand{\KvzipRecTwoFive}{0.51}
\newcommand{\KvzipRecFifty}{0.71}

\newcommand{\KvzipPairTwoFive}{$+0.34$ [$+0.27$, $+0.42$]}
\newcommand{\KvzipPairFifty}{$+0.21$ [$+0.15$, $+0.28$]}
\newcommand{\KvzipDenseAgree}{634}
\newcommand{\KvzipDenseTotal}{658}
\newcommand{\KvzipIdentSame}{8}
\newcommand{\KvzipIdentTotal}{9}
\newcommand{\CrossBothN}{227}
\newcommand{\KvzipScoreSecs}{0.8}
\newcommand{\CrossRows}{160}
\newcommand{\CrossAgree}{150}
\newcommand{\CrossNetBias}{+0.04}
\newcommand{\KvzipMkN}{59}
\newcommand{\KvzipMkTen}{0.92}
\newcommand{\RouterMkTen}{0.93}
\newcommand{\KvzipMkTwoFive}{0.95}
\newcommand{\RouterMkTwoFive}{0.95}
\newcommand{\KvzipVtN}{55}
\newcommand{\KvzipVtTen}{0.96}
\newcommand{\RouterVtTen}{1.00}
\newcommand{\KvzipVtTwoFive}{0.98}
\newcommand{\RouterVtTwoFive}{1.00}
\newcommand{\QasperRefB}{0.83}
\newcommand{\QasperHtwoB}{0.32}
\newcommand{\QasperRandB}{0.38}
\newcommand{\QasperBaseRcB}{0.48}
\newcommand{\QasperSnapB}{0.75}
\newcommand{\QasperPairRcB}{0.55}
\newcommand{\QasperPairRaB}{0.36}
\newcommand{\QasperPairAnnB}{0.59}
\newcommand{\QasperPairAnnBfifty}{0.71}
\newcommand{\QasperLLTen}{0.35}
\newcommand{\QasperLLTwentyFive}{0.50}
\newcommand{\QasperLLFifty}{0.63}
\newcommand{\QasperPairRcSeedLo}{0.51}
\newcommand{\QasperPairRcSeedHi}{0.55}
\newcommand{\QasperPairRaSeedLo}{0.36}
\newcommand{\QasperPairRaSeedHi}{0.37}
\newcommand{\QasperStrictRc}{0.56}
\newcommand{\QasperStrictRa}{0.38}

\usepackage[preprint]{icml2026}

\icmltitlerunning{Causal Evidence Sets for Global Context Pruning}

\begin{document}

\twocolumn[
\icmltitle{Learning What Matters: Causal Evidence Sets for Global Context Pruning}

\begin{icmlauthorlist}
\icmlauthor{James E. Allchin}{ind}
\icmlaffiliation{ind}{Independent Researcher}
\icmlcorrespondingauthor{James E. Allchin}{jim@allchin.com}
\end{icmlauthorlist}

\icmlkeywords{long context, causal intervention, attention, pruning}

\vskip 0.3in
]

\printAffiliationsAndNotice{}

\begin{abstract}
Pruning a long context means committing to the blocks a model will keep, and a common selector is distilled from a dense teacher's attention. That assumes attention shows which context the answer depends on. We test the assumption on retrieval tasks where the evidence is known exactly, by masking context and measuring whether the answer changes. Attention and causal dependence disagree. Teachers attend to outdated facts the answer does not use, and a selector distilled from that attention routes a multi-hop task at 36\% to 98\% across runs. The same selector trained on causal evidence sets reaches 99\% or better on every run, and known evidence labels beat attention-imitation and accumulated-attention eviction through 32B when the context must be pruned before the question arrives. A reconstruction compressor, KVzip, is stronger on natural-text QA with native per-head KV eviction. Under our paragraph-level adaptation, its scores do no better than labels recovered by intervention.
\end{abstract}

\section{Introduction}

Sparse attention reduces computation by letting each query read a subset
of the context. Some methods train a small \emph{selector} network to
imitate a dense \emph{teacher}'s attention
\citep{ahmad2026spotattention}. This assumes the teacher attends to the
context its answer depends on.

\looseness=-1
This paper tests that assumption in a setting where the context a model
needs is known independently of its attention. Our tasks are synthetic
key-value retrieval problems. The context is a sequence of fixed width
\emph{blocks}, each holding one \emph{record}, followed by a query. A
record is a single stored fact, a key and the value filed under it. \emph{Single-record} tasks have one
record determining the answer. In \emph{multi-hop} tasks the queried
record points to a second record, that record to a third, and the third
holds the value. \emph{Latest-write} tasks write the same key several
times, and only the last value is correct. By construction we know
which records determine the answer. A set of blocks is \emph{sufficient}
for an example if restricting the model's attention to those blocks and
the query still yields the correct answer. Sufficiency is verified by
intervention, masking attention into a block at every layer with token
positions unchanged. The supervision target is the \emph{causal evidence set}, the
annotated task evidence whose sufficiency the interventions
verify (Section~\ref{sec:labels}). Causal means dependence under
masking, not a recovered computation graph, and the label is not always
a minimal sufficient set. On these tasks we train dense teachers from several random seeds,
freeze them, and compare two supervision targets for a selector. We call
that selector a \emph{router} when it chooses one block set for the whole
forward pass, the protocol of KV eviction and prompt compression
\citep{zhang2023h2o, li2024snapkv}. Deploying a router masks every block
it did not select, so the model answers from the kept blocks alone. The
resulting accuracy is its \emph{routed accuracy}.

\begin{figure}[t]
\centering
\includegraphics[width=\linewidth]{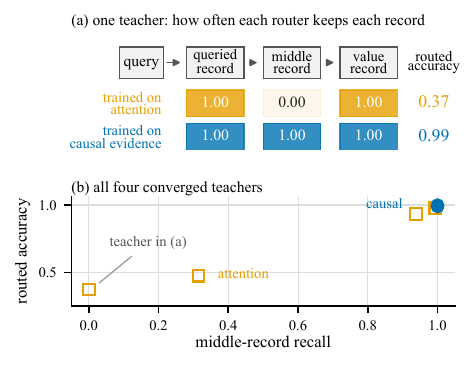}
\caption{Multi-hop routing at the 10\% block budget (3 of 32 blocks). (a) Keep rate of
each chain record, on the teacher where the supervisions differ most.
(b) Routed accuracy against middle-record recall, one mark per
teacher. Both panels average three router seeds.}
\label{fig:chain}
\end{figure}

\looseness=-1
Dense teachers solve multi-hop at 98\% or better. The pointer steps are
resolved in earlier layers at earlier positions, and how much of the
chain the answer-position attention covers depends on the training run.
On some training runs, a router distilled from that attention keeps
the two ends of the chain and misses the middle record
(Figure~\ref{fig:chain}). With identical architecture, data, and
budget, it routes at 0.36 to 0.98 across four converged teachers.
Dense accuracy does not tell those teachers apart. The same router
trained on causal evidence sets keeps all three records and routes
at 0.99 to 1.00. Attention shows where the answer is read, not
where it was computed. Checking the target requires interventions.

The single-record tasks give the same conclusion in a milder form. On
latest write, every teacher, at every seed, spends attention on the
obsolete writes on every example. That attention is spread across many
blocks (Section~\ref{sec:stale}). Evidence sets measured by the same
interventions agree across seeds about twice as well as adaptive
attention. Planting redundant copies of the answering record breaks the
single-block interventions, since masking one copy changes nothing. The
collection of interchangeable sufficient sets survives, and agrees better as
redundancy grows (Section~\ref{sec:stability}).

We train routers with identical architecture and inputs, one imitating
block-level attention distributions (the imitation router), one trained
on causal evidence sets (the causal router), each deployed as a hard mask
on the frozen teachers at a 10\% block budget.

The paper makes four contributions.
\begin{itemize}
\item A testbed with known evidence. Interventions show that attention
covers obsolete writes, misses multi-hop chains at the answer, and
varies across seeds whose evidence stays fixed. Annotated evidence sets
do not.
\item Matched routing experiments. The label determines routed
accuracy, independent of the loss. The router must also be able to
represent the selection.
Per-query selection lets attention routing succeed while pruning
nothing. A backward trace through attention supervises these
teachers' chains as well as annotated sets. Interventions still separate
current from obsolete evidence, and they remain stable under redundancy.
\item An annotation-free estimator of the causal evidence set. On the
constructed tasks, routers trained on its labels nearly match
annotation-trained ones. Every single-signal alternative fails where
the measurements predict.
\item The same measurements on frozen pretrained models across three
families. Attention prefers obsolete records to current ones. Keeping the
current record repairs conflicting facts. Label recovery stays exact on the pretrained single-record tasks.
Causal supervision leads imitation under length shift and on
counterfactual SQuAD. On long-context eviction the same labels beat
accumulated-attention policies through 32B. On a multi-hop task the models solve, attention pooled over heads
misses the chain, single heads selected on held-out examples, with or without
annotations, recover most of it, and no tested router does.
\end{itemize}

Some learned selectors imitate teacher attention
\citep{ahmad2026spotattention, gao2024seerattention}; other work questions
attention as importance or replaces it with reconstruction or
intervention signals (Section~\ref{sec:related}). To our knowledge none
tests attention as a global pruning target against known evidence. In
end-to-end sparse training the model co-adapts to the mask until
learned gates barely beat random ones
\citep{aquinomichaels2026routing}; we therefore train teachers dense and
freeze them.

\section{Setup}

\subsection{Tasks}
\label{sec:tasks}

Every example is a sequence of $n$ blocks of $w$ tokens followed by a short
query, with $n{=}32$ and $w{=}8$ unless stated. Each block holds one record,
\texttt{WRITE} $f$ $m$ $v$, padded with filler: a two-token key (family $f$,
member $m$) and a one-token value $v$. Two-token keys give graded distractor
similarity, since a distractor can share the family token without sharing
the key. The query is \texttt{QUERY} $f$ $m$ $v$, and the model is scored on
predicting $v$ at its position.

\begin{table*}[t]
\caption{The five tasks, with minimal task-evidence sets known by
construction.}
\label{tab:tasks}
\begin{center}\small
\begin{tabular}{lll}
\toprule
Task & Answer determined by & Minimal task-evidence sets \\
\midrule
Unique retrieval & the one record with the queried key & one single block \\
Latest write & the last of several writes to the key & one single block \\
Duplicate evidence & any of $m$ identical records & $m$ single blocks \\
Multi-hop (2 hops) & a chain of two pointers and a value & one set of 3 blocks \\
Aggregation & the count of records in a family & one set of all marked blocks \\
\bottomrule
\end{tabular}
\end{center}
\end{table*}

Table~\ref{tab:tasks} names the five tasks. In latest write, earlier writes
carry different values. Each duplicate alone is sufficient. We call the
set of minimal task-evidence sets of an example its \emph{family}. In
multi-hop, a pointer record holds a second key in place of a value.
Aggregation requires all marked blocks and is the intended boundary case.

The value is predicted directly at the query key. An answer marker
between key and value prevents the retrieval circuit from forming at this
scale. Training sequences carry eight queries each; evaluation is
single-query, an exact prefix of the training format
(Appendix~\ref{app:protocol}; \citealp{arora2023zoology}).

\subsection{Teachers, interventions, and sanity checks}
\label{sec:teachers}

Teachers are decoder-only transformers, 6 layers, 8 heads, model width 384,
10.7M parameters, rotary position embeddings, trained with AdamW to high
accuracy on a task mixture (retrieval \TeacherRetrievalRange{},
aggregation \TeacherAggRange{}). Multi-hop reaches
\TeacherMultihopRange{} in four of five seeds. Seed 3 reaches
\TeacherMultihopFailed{} and is excluded from multi-hop experiments.
Per-seed finals are in the repository's results files. All teachers are then frozen.
No sparsity is imposed during training, so the supervision labels cannot
be absorbed into the weights \citep{aquinomichaels2026routing}.

Masking a block adds $-10^4$ to the attention logits of its positions at
every layer for every query outside the block, the same fill value the
causal mask uses. Token positions do not move, so the intervention
removes information, not positional structure. Restricting to a set means
masking its complement. Both directions are checked. For solved examples,
keeping only the annotated sufficient set preserves the answer, and
masking the annotated block flips it on 98\% of unique-retrieval examples
while masking a random other block flips it on 0\%. Masking and
overwriting a record with filler agree on 100\% of 200 checked retrieval
examples. On unique retrieval, the top attended block at the answer
position is the annotated block on 99\% of solved examples, so where one
block matters and no distractor shares the key, attention and sufficiency
coincide.

\subsection{Labels}
\label{sec:labels}

\emph{Task evidence} is the annotated records that determine the
answer by construction. A block is \emph{necessary} if masking it changes
the full-context answer. A set is \emph{sufficient} if restriction to that set
preserves the answer, as defined above. On solved single-record examples
these three coincide. Sufficiency and the masking effect score the gold
answer token, not the teacher's emission. Examples the teacher answers
incorrectly still receive labels measuring support for the correct
answer. Necessity is an exception. It compares against the teacher's own
full-context prediction. Redundancy breaks necessity
(Section~\ref{sec:stability}). If the rest of a chain is hidden, the model
often copies the only remaining value, so a sufficient set can be smaller
than the task evidence (Appendix~\ref{app:minimality}).

Attention labels take the $k$ blocks with the largest attention mass at
the answer position, pooled over layers and max or mean over heads, plus
an adaptive variant that takes the smallest set covering 90\% of that
mass over context blocks. Ablation ranking takes the $k$ blocks whose
individual masking most reduces the answer log probability.

Routing trains on the annotated task-evidence family, the causal
evidence set, whose sufficiency the interventions verify. Stability and
agreement use a \emph{measured evidence set} that is purely behavioral.
If some single block preserves the answer on its own, the set is every
block that does so, recovering all copies under redundancy. If no single
block suffices, the set is every block whose masking flips the answer.
On chains that behavioral set can undershoot the annotation, because a
lone endpoint lets the model copy the remaining value
(Appendix~\ref{app:minimality}). Section~\ref{sec:recovered} recovers an
estimate of the annotated family from interventions alone. The measured
set has no fixed size; comparisons against attention therefore use the
adaptive variant, which also chooses its own size.

\section{Attention includes unused evidence}
\label{sec:stale}

On the latest-write task, where a key is written several times and only
the final write is correct, the adaptive attention set includes at least
one obsolete write on 100\% of examples, for every one of five seeds.
Attention retrieves everything that matches the queried key and later
layers decide which write is current. At a $k{=}3$ budget
(10\% of 32 blocks, rounded down to 3),
attention labels agree across seeds about as well as measured evidence
sets do (Jaccard set overlap 0.76). Stability alone does not certify a
target whose stable sets contain evidence that does not matter.

That attention is also spread out. Covering 90\% of the answer-position
attention takes 13 to 15 of 32 blocks depending on seed, against 1.4 to 1.6
blocks for the measured evidence set, so a router distilled from attention
learns a nine-times-larger target than the evidence requires.

\section{Causal evidence sets are stable}
\label{sec:stability}

On the single-record tasks the measured evidence set is about twice as
consistent across seeds as adaptive attention
(Table~\ref{tab:agreement}, appendix). Different training runs solve these
tasks with different attention and the same evidence. Redundancy separates
the estimators.
With $m$ copies planted, masking any one copy stops changing the answer,
so single-block flip rates collapse and ablation-ranked labels decay with
them, while the family label grows more consistent
(Figure~\ref{fig:redundancy}, appendix). When several answers are equally
good, which one a model uses cannot be recovered from the answer, but the
set of usable answers can, and it is the stable object. The measured set
is accurate against construction. The measured family equals the
annotated one on 70 to 82\% of examples and covers a correct sufficient set on more
than 99\%.

The instability carries into the routers that learn from these labels.
Training a router per label type, per source teacher, and per initialization, then
scoring each on all five teachers, separates the source of the labels from
the initialization the router started at
(Table~\ref{tab:app-transfer}). In every
condition, a destination's score varies two to six times more with the
source of its attention labels than with its initialization, the source
matters less than the initialization for causal evidence sets, and the
mean off-diagonal attention score exceeds the own-teacher diagonal, while
the causal means differ by at most 0.004. Dense accuracy
does not say which teacher supplies usable attention labels.

\section{Causal supervision improves routing}
\label{sec:routers}

A router scores context blocks given the token sequence, with the
supervised position removed, since the answer is never available at
routing time. It never reads teacher activations, so one trained router
runs unchanged on any teacher. The \emph{pooling router} embeds tokens
with its own table, mean-pools each block and the query, and scores each
block independently from its summary, the query summary, and relative
position. The \emph{chain router} replaces mean pooling with a projection
per within-block offset, so a pointer's source and destination keys stay
distinguishable, and adds two transformer layers over the summaries, so a
block's score may depend on other blocks. At evaluation the router's top
$k{=}\max(1, \lfloor n/10 \rfloor)$ blocks stay visible and everything else is
masked, using the operator of Section~\ref{sec:teachers}. The query is
always visible. Both routers are small: 23K and 111K parameters, under
0.3 ms per 32-sequence batch, flat from 32 to 128 blocks.

All routers train on the same example stream from the same label teacher.
The imitation router distills the teacher's attention at block
granularity: it minimizes the Kullback--Leibler (KL) divergence between
the teacher's attention over blocks at the answer position and the
router's block distribution. This is the recipe of
\citet{ahmad2026spotattention} up to their per-query scope, and their
top-$k$-restricted objective changes nothing
(Table~\ref{tab:ablation}, appendix).
The causal supervision trains on causal evidence sets with a coverage
loss. Per example it applies binary cross entropy against each acceptable
set's indicator, then combines those losses with a softmin (a smooth
minimum) that anneals to a hard minimum over training. Covering any one
sufficient alternative is enough, so the router keeps one copy instead of
all of them. A random-gate control keeps $k$
random blocks \citep{aquinomichaels2026routing}. Routed accuracies use 300 to 500 examples, so a single
cell carries a 95\% binomial interval of $\pm 0.06$ midrange and $\pm 0.02$
near the extremes. Differences between the two routers are paired within a teacher and
reported with the standard error over teacher seeds, which is the
uncertainty we interpret them against.

In distribution the supervisions stay within three points of each other
and track the dense teacher (Table~\ref{tab:routing}). On unique retrieval both sit slightly above it,
since masking distractors removes interference, a repair effect that
returns at scale in Section~\ref{sec:pretrained}. A stale-4 reversal in a single-router run vanishes under router-seed
averaging (Table~\ref{tab:routing}).
Under length shift they separate on unique retrieval
(Figure~\ref{fig:gap}, appendix). At four times
the training length the causal router leads on every seed, by \UniqueFourGapPts{} points on
average, and by 8 to 13 points after retuning the imitation router
(Table~\ref{tab:app-guard}, appendix). Own-teacher training leaves the gap larger in all nine conditions
(Table~\ref{tab:app-own-teacher}), so it is not a transfer penalty.
On latest write the supervisions stay within a point through doubled
length, and a small causal lead emerges at 4$\times$
(Table~\ref{tab:routing}), matching the replication's direction. The
teacher's attention concentrates on the write records, so its top blocks
and the annotated set largely agree. Dense teachers themselves fail at 4$\times$; the 2$\times$ rows show
the gap where dense still functions.

% Generated by analysis/make_main_tables.py from runs/. Do not edit by hand.
\begin{table*}[t]
\caption{Single-record routed accuracy at the 10\% block budget
(pooling routers, seed-0 labels; mean over five teachers and five
router seeds). Stale-4 evaluates latest write with four obsolete writes; training uses two. $\Delta$ is causal minus
imitation, paired $\pm$ SE. Bold marks a 5\% paired $t$. A dash rounds
to zero.}
\label{tab:routing}
\centering\footnotesize
\begin{tabular}{lccccc}
\toprule
Condition & Dense & Imitation & Causal & $\Delta$ & Random \\
\midrule
Unique retrieval & 0.93 & 0.96 & \textbf{0.99} & $+0.028 \pm 0.005$ & 0.09 \\
Latest write & 0.97 & 0.93 & 0.93 & -- & 0.07 \\
Duplicate evidence & 0.99 & 0.99 & \textbf{1.00} & $+0.004 \pm 0.001$ & 0.16 \\
Latest write, stale-4 & 0.96 & 0.93 & 0.93 & $-0.001 \pm 0.001$ & 0.08 \\
Unique, family distractors & 0.94 & 0.97 & \textbf{0.99} & $+0.022 \pm 0.007$ & 0.09 \\
Unique, 2$\times$ length & 0.78 & 0.89 & \textbf{0.94} & $+0.047 \pm 0.015$ & 0.09 \\
Latest, 2$\times$ length & 0.74 & 0.90 & 0.90 & $-0.000 \pm 0.007$ & 0.10 \\
Unique, 4$\times$ length & 0.32 & 0.62 & \textbf{0.73} & $+0.110 \pm 0.012$ & 0.10 \\
Latest, 4$\times$ length & 0.42 & 0.77 & \textbf{0.80} & $+0.029 \pm 0.008$ & 0.09 \\
\bottomrule
\end{tabular}
\end{table*}

Multi-hop separates the supervisions, and needs both the right target
and a router that can express the selection (Table~\ref{tab:chain}). Routing on labels
directly, with no learned router, separates the targets. The annotated
chain and the ablation ranking score 0.98 to 1.00 on all four seeds; the
teacher's top attended blocks range from 0.41 to 0.98 depending on seed
and head pooling, and the best single layer's attention never exceeds
0.55. Pooling routers fail under every supervision
(0.01 to 0.10, Table~\ref{tab:ablation}, appendix). Mean pooling collapses a pointer's
source and destination keys into one summary, and no loss can teach what the
encoding cannot represent. The chain router removes that constraint, and the supervisions
separate. Trained on annotated chains it routes at 0.99 to 1.00 on every
converged teacher. Trained by imitation it ranges from 0.36 to 0.98
(Table~\ref{tab:chain-per-teacher}). The spread is not router noise: router seeds within a teacher agree to
within 0.08, while across teachers imitation moves by sixty
points. It tracks how often the router selects the chain's middle block
(Figure~\ref{fig:chain}b). Causal supervision keeps all three
blocks.
Attention at the answer position is not a reliable routing target on this
task.

This comparison keeps one set for the whole pass, the protocol in
which context is pruned. Selecting per query and per layer instead, as
deployed block-sparse attention does, attention routes at 0.94 to 1.00
while keeping the entire context resident
(Appendix~\ref{app:protocol-scope}). Per-query selection restricts what
each query reads but prunes nothing.

% Generated by analysis/make_main_tables.py from runs/. Do not edit by hand.
\begin{table*}[t]
\caption{Multi-hop routing at $k{=}3$. Oracles select from labels;
ranges are the four converged seeds. Chain recall is first
pointer/second pointer/value. Best layer is post hoc, an upper bound.}
\label{tab:chain}
\centering\footnotesize
\begin{tabular}{llcc}
\toprule
Selector & Supervision & Routed & Chain recall \\
\midrule
Dense teacher (no routing) & & 0.99 & \\
Oracle on labels & annotated chain & 0.99--1.00 & \\
Oracle on labels & ablation ranking & 0.98--0.99 & \\
Oracle on labels & teacher attention & 0.41--0.98 & \\
Oracle on labels & attention, best layer & 0.21--0.55 & \\
Chain router, seed 0 & attention (KL) & 0.37 & 1.00/0.00/1.00 \\
Chain router, seed 0 & causal (coverage) & \textbf{0.99} & 1.00/1.00/1.00 \\
Chain router, all teachers & attention (KL) & 0.36--0.98 & \\
Chain router, all teachers & backward trace (coverage) & 0.98--0.99 & \\
Chain router, all teachers & causal (coverage) & \textbf{0.99--1.00} & \\
Random gates & & 0.04 & \\
\bottomrule
\end{tabular}
\end{table*}

\looseness=-1
The gaps could still come from the loss, or from attention elsewhere
covering the chain when the answer position does not. Tables~\ref{tab:ablation} and~\ref{tab:chain-cross} (appendix) test
both, on pooling routers and on the chain router, by crossing label
type with loss form and adding an attention label pooled over every
position and layer. Each label scores about the same
under either loss, and causal evidence leads under both: the failure
follows the label. All-position
attention is the worst label in the table, at or below random. Attention mass summed over positions is mostly local structure, so
averaging more of it buries the retrieval signal. The chain is present in the attention weights.
Composing them across layers with a backward max-product trace from the
answer (the \emph{backward trace}) routes at 0.90 to 0.99. Ranking blocks
by answer-position attention (the answer-position \emph{readout}) gets
0.41 to 0.87 (Appendix~\ref{app:readouts}). A router
trained on the backward trace matches annotated sets on chains, 0.98 to
0.99 across teachers
(Table~\ref{tab:chain}), at one forward pass per example against two per
block for the intervention labels. The repair has two bounds. The backward
trace keeps an obsolete write where annotated sets keep none, and it does
not recover the pretrained multi-hop chain on frozen models
(Section~\ref{sec:pretrained}).

The teacher's training problem also changes whether attention is a usable
target. We harden multi-hop so that nothing short of resolving the chain
answers the query. Pointer and value records are shuffled in position, so
chain role is not a location, and each context holds two distractor
pointer chains. A teacher trained on this variant must attend the chain.
Imitation then recovers: 0.97 against 0.99 to 1.00 for causal
supervision. A second teacher learned the other tasks but not hardened
multi-hop (0.05 dense). It has the chain circuit but cannot shield it
from the distractors. The router trained on the first teacher transfers
zero-shot and routes the second at 0.91. On three-hop chains neither
router nor teacher ever saw, selection stays essentially perfect
(full-chain recall 0.997 to 1.0) while routed accuracy reaches 0.38
against 0.05 dense. The router finds the right blocks. The teacher cannot
compute three hops.

\section{Labels need no evidence annotations}
\label{sec:recovered}

\looseness=-1
We label 8{,}000 training examples by intervention and train routers on
these labels, using annotations only to score recovery. The teacher answers
\RecoveryStreamWrongPct{} of these examples incorrectly; they are kept
and labeled by the same interventions against the gold token.

Each single-signal estimator fails where the measurements say it must
(Table~\ref{tab:recovered}, appendix). Sufficiency alone undershoots on
chains, where a fallback keeps the restricted answer alive. Ranking the
$k$ blocks whose masking most reduces the answer log probability solves
chains and fails under redundancy, where masking any one copy leaves the
answer unchanged (Section~\ref{sec:stability}). On latest write one block
suffices, so leftover slots are irrelevant.

The regimes separate in the interventions themselves. Masking a block the
computation runs through moves the answer log probability by a nat or
more. Masking a redundant copy or a filler block has negligible effect.
The estimator therefore keeps the ranked blocks that clear a 0.25-nat
floor, capped at the routing budget. Where nothing clears that floor, it
keeps every single block that preserves the answer on its own. That
order is the reverse of the measured set in Section~\ref{sec:labels}.
Effect-first ranking estimates the annotated family. Singleton-first
branching adopts the copy fallback on chains (pseudocode in
Appendix~\ref{app:recovered}).

Its labels match the annotated union on 87\% of examples and contain a
sufficient set on 92\%. Routers trained on them nearly match
annotation-trained routers (Tables~\ref{tab:routing} and \ref{tab:chain}).
The router generalizes past its labels: 82\% of its chain labels are
complete, yet routed chains exceed 98\%. An attribution baseline at the
same budget, labeling the top blocks by input-gradient saliency, lands
between the targets on every task (Table~\ref{tab:recovered}). Dependence
signals beat attention even without interventions. The intervention
labels stay ahead everywhere.

\section{Transfer to pretrained models}
\label{sec:pretrained}

\looseness=-1
The synthetic teachers could be a special case, so we ask the same
question on frozen Qwen2.5-Instruct models. The tasks are natural language,
one record per sentence (``The code for NARO is 4821.''), a question at
the end, and later sentences that update earlier ones for conflicting
facts. Record spans come from tokenizer offsets. Masking uses the same
operator as the synthetic teachers. Scoring is forced choice among
written candidates, so a mask cannot be bypassed by generating.

Table~\ref{tab:pretrained} shows the synthetic structure repeated.
Masking the
current record flips the forced choice to an obsolete value, masking a random
record does not, and the current record alone is sufficient. On unique
retrieval every model is perfect dense, and with the target record masked
the choice among four candidates falls to the chance flip rate. Candidates
come from outside the context, since in-context ones allow answering
by elimination. The attention finding persists where the task is solved.
The 3B and 7B models answer nearly every conflicting-fact query while
giving the obsolete record more attention than the current one on about
half of them, and the models below 3B are less reliable. Scale fixes the behavior before it fixes the attention.

We train a record-level router against the frozen 3B model, scoring
records instead of blocks, with a cosine record-question similarity for
rare name tokens. The causal router
trains on the annotated record, the imitation router on answer-position
attention over records, both on 2{,}000 examples, deployed by keeping the
top two records. The causal router matches the dense model on both tasks
and stays there when the record count doubles past training. The imitation
router falls to its random control at the doubled count, and the 7B model
repeats the pattern (Appendix~\ref{app:sink}). Most of that deficit is the
first-token attention sink \citep{xiao2024streamingllm,
gu2025attentionsink}: answer-position attention puts its largest
record mass on the first record on every training example, so the imitation
target spends one of two budget slots there. Dropping that record from the
target and the selection removes the collapse on conflicting facts, where
sink-free imitation equals the causal router at both record counts on both
models. On unique retrieval a gap survives the matched correction on the Qwen
models, 6 to 26 points, widening with record count as in
Section~\ref{sec:routers} and again as the budget tightens. At one record the correction narrows but does not close the gap
(Table~\ref{tab:budget-sink}; Appendix~\ref{app:sink}).

\looseness=-1
Yi-1.5-9B solves both tasks dense, gives obsolete records more attention than
current ones on 72\% of examples, and routes the same way. Imitation
degrades on Yi. It does not collapse, so the routing signal varies by
family, and the sink correction reproduces on it
(Table~\ref{tab:budget-sink}). Gemma-2-9B fails the other way around.
Dense, it scores 0.56 on conflicting facts, and the causal router keeping
two records lifts it to 0.98 or better, because masking them removes a
distraction the model cannot resolve. Across three model
families, from 0.5B to 9B, attention puts mass on obsolete evidence the answer does not depend
on, and causal evidence sets do not; some models discount that
evidence, others are harmed by it.
The conflicting-fact repair also holds under a position-preserving
content overwrite (Appendix~\ref{app:overwrite}).

% Generated by analysis/make_main_tables.py from runs/. Do not edit by hand.
\begin{table*}[t]
\caption{Frozen pretrained models on conflicting facts (400 examples,
16 records, 2 obsolete writes; flip and sufficiency on solved
examples). Obsolete $>$ current: the obsolete record out-attends the
current one; the no-sink control repeats it
(Table~\ref{tab:trace-pretrained}).}
\label{tab:pretrained}
\centering\footnotesize\setlength{\tabcolsep}{3.5pt}
\begin{tabular}{lcccccc}
\toprule
Model & Accuracy & Flip (causal) & Flip (random) & Sufficient alone & Obsolete $>$ current & no sink \\
\midrule
Qwen2.5-0.5B & 0.23 & 1.00 & 0.29 & 1.00 & 0.83 & -- \\
Qwen2.5-1.5B & 0.60 & 1.00 & 0.03 & 1.00 & 0.72 & -- \\
Qwen2.5-3B & 0.99 & 1.00 & 0.00 & 1.00 & 0.58 & 0.48 \\
Qwen2.5-7B & 0.96 & 1.00 & 0.01 & 1.00 & 0.43 & 0.29 \\
Yi-1.5-9B & 1.00 & 1.00 & 0.01 & 1.00 & 0.72 & 0.65 \\
Gemma-2-9B & 0.56 & 1.00 & 0.05 & 1.00 & 0.45 & -- \\
\bottomrule
\end{tabular}
\end{table*}

The multi-hop omission transfers in a stronger form. On a natural-language
multi-hop task whose answer record never names the queried key, both
records are necessary for the Qwen models. Yi can sometimes still answer
after the pointer is masked, above the rate of guessing a random pair.
All three models solve the task dense (81 to 95\%)
while no readout pooled over heads finds the chain
(Table~\ref{tab:two-hop-readouts}). At the chain budget, answer-position attention, rollout, and the
backward trace keep both chain records on at most a tenth of solved
examples and route at 0.21 to 0.67, against 1.00 for the annotated pair
on the same examples (Table~\ref{tab:two-hop-readouts}, appendix). The
trace that repairs the synthetic chains does not repair these; it
recovers chains the teacher resolves in attention. Pooling hides the
chain. A single head, selected on held-out examples for keeping a chain
record on top across record orderings \citep{wu2026expertheads}, keeps
both records on \ExpertBothLo{} to \ExpertBothHi{} of solved examples and
routes at \ExpertRoutedLo{} to \ExpertRoutedHi{}
(Table~\ref{tab:expert-heads}, appendix). Selecting it from intervention
labels instead of the annotation picks the same head on all three models. Learned
routers fail here too. Routed
accuracy stays at the random control under every label source, the
annotated pair included, though annotated labels lift full-pair recall to
0.07 to 0.13 against 0.01 random, far too weak to move answers. For the
7B model, neither a cross-record mixer nor 20{,}000 annotated examples
helps (Table~\ref{tab:two-hop-router}).

The recovery principle of Section~\ref{sec:recovered} transfers at lower
cost. In this single-record regime necessity alone identifies the record.
Hiding annotations on the 3B model, one masked forward per record recovers
them exactly, and a router trained on those labels matches the
annotation-trained router. Recovery stays exact to 128 records
(Table~\ref{tab:scaling}). The cost is one clean pass plus one masked pass per record per
example: \CostSixteenMinutes{} minutes for 2{,}000 examples at 16
records on one consumer card and \CostTwelveEightMinutes{} minutes for
\CostTwelveEightExamples{} examples at 128, about \CostMillionDays{} days
per million examples at 16 records and \CostMillionDaysTwelveEight{} at 128, a one-time cost
amortized over deployment. The procedure needs a scoreable answer, not
task structure, so it extends to natural data. On counterfactual SQuAD
(\citealp{rajpurkar2016squad}; 16-sentence contexts, with substitute answers so parametric memory
cannot solve), masking recovers the gold evidence sentence on 92\% of
examples, and at a two-sentence budget the recovered-label router beats
attention imitation, 0.47 against 0.35 with dense 0.64 and a
gold-evidence oracle at 0.77. Both routers sit well below dense,
keeping the evidence on 59\% and 41\% of examples, so this is
evidence about labels, not a working QA router.

\begin{figure}[tb]
\begin{center}
\includegraphics[width=\linewidth]{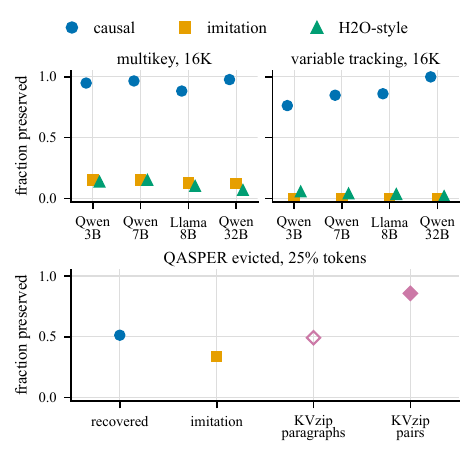}
\end{center}
\caption{Pre-query eviction at 16K (10\% of blocks) and physical
eviction on QASPER (25\% of tokens).}
\label{fig:eviction-scale}
\end{figure}

\looseness=-1
We also test the same label choice on long-context eviction, where the
cache must shrink before the question exists. On RULER retrieval and
tracking, a block router trained at 4K on the generator's evidence labels keeps
the evidence a not-yet-asked question will need. The matched imitation
router drops to chance, and accumulated-attention eviction keeps the
obsolete write. Appendix~\ref{app:external} evaluates three protocols,
masking at prefill with and without the question, and dense prefill
followed by physical eviction; the ordering is the same under all
three. Figure~\ref{fig:eviction-scale} shows the ordering replicating across
scoring models, 0.98 against 0.07 for H2O-style eviction on Qwen-32B multikey at a
10\% budget, with the largest margins at 32B, and the advantage
survives physical cache eviction after a dense prefill
(Table~\ref{tab:evict}, appendix). On QASPER, where questions are reader paraphrases, annotated evidence
fits a 25\% budget and attention-based eviction scores below random.
A selector over frozen 0.5B paragraph encodings, trained on recovered
intervention labels and scored by 7B answer preservation, leads its
matched attention twin at every budget on the dense-correct questions of
\QasperDocs{} held-out documents (\QasperPairRcB{} against
\QasperPairRaB{} at 25\%; annotated-evidence reference \QasperRefB{};
Table~\ref{tab:qasper-pair}, appendix). It is competitive with
LLMLingua-2 token compression.

\paragraph{External query-agnostic compression.} Native KVzip
\citep{kim2025kvzip} preserves more QASPER answers under physical
eviction than the recovered-label selector
(Table~\ref{tab:kvzip}, appendix). On the 16K conditions the two have
similar point estimates and no resolved difference. Evidence supervision
repairs the tested imitation failures. It is not the strongest compressor
on this natural-text evaluation. KVzip's scores at paragraph granularity
preserve \GranDirectTwoFive{} at 25\%. As labels for the same selector
they trail recovered labels
(Tables~\ref{tab:kvzip-granularity} and~\ref{tab:qasper-kvzip-labels},
appendix).

\section{Limits of sparse routing}
\label{sec:limits}

\looseness=-1
Aggregation is the case where sparse routing fails. An
oracle keeping the annotated sufficient blocks first is the
annotated-evidence reference; no tested router exceeds it
(Figure~\ref{fig:budget}, appendix). Unique retrieval is flat
from $k{=}1$. Multi-hop steps to 0.99 at the chain size. Aggregation
stays at chance through $k{=}16$. The count fails even if the mask
leaves every marked block. Where computation is distributed, the
task's and the model's sufficient sets diverge, and no tested
selection routes under this mask.

\section{Related work}
\label{sec:related}

\looseness=-1
\textbf{Sparse attention.} Training-free methods select or evict context
by the model's own attention or key statistics \citep{zhang2023h2o,
li2024snapkv, tang2024quest, jiang2024minference}; learned selection
trains the signal instead \citep{yuan2025nsa, lu2025moba,
gao2024seerattention, ahmad2026spotattention}, and content-based routing
clusters or hashes queries and keys \citep{roy2021routing,
kitaev2020reformer}. Earlier sparse transformers fixed the pattern
\citep{beltagy2020longformer, zaheer2020bigbird}. When gates train end to
end, co-adaptation confounds them \citep{aquinomichaels2026routing}.
\citet{an2026restkv} evict by layer-wise output reconstruction,
agreeing that attention misses the effect of removing a token; we test
their indicator in our regime (Table~\ref{tab:restkv}, appendix) and
make no latency claim. \citet{hwang2026bpkv} score evictions by an
estimated change in the model's next-token distribution. Query-agnostic eviction has
its own line. KVzip
scores entries by reconstructing the context from the cache
\citep{kim2025kvzip}. KVzap, Fast KVzip, and EchoPress approximate that
score more cheaply \citep{jegou2026kvzap, kim2026fastkvzip,
lin2026echopress}. Others rank
by future rewards or predicted future attention
\citep{moschella2026kvpolicy, devoto2025expectedattention}, or condition
on a task description \citep{corallo2026cachenotes, park2026taskpress}.
Our selector instead learns from per-example evidence labels. Theory
separates compressors that know the query distribution from those that
do not \citep{haverbeck2026risk}. We measure KVzip; it leads on natural text
(Section~\ref{sec:pretrained}), so our pre-query results bound the
passive-prefill readouts we test, not that family.
\citet{guo2025leaf} intervene to drop confounding tokens when
distilling a student; our labels mark what a router keeps.
\citet{xiao2024streamingllm} keep the
attention-sink tokens because evicting them destabilizes generation. A
distilled target has the opposite problem: it spends selection budget on
them.

\textbf{Attention as importance.} Whether attention weights are reliable
importance estimates has been debated for classification models
\citep{jain2019attention, serrano2019attention, wiegreffe2019attention,
abnar2020quantifying}. Those studies compare attention with other
estimates on natural data. We score it as a pruning target against
evidence fixed by construction.

\textbf{Sufficient subsets and rationales.} \citet{carter2019sis} find input subsets that alone
preserve a classifier's decision and collect disjoint ones when several
exist, and ERASER scores annotated rationales by sufficiency and
comprehensiveness \citep{deyoung2020eraser}. Inputs reduced until only the
prediction survives can lose the evidence a reader would use
\citep{feng2018pathologies}, the shortcut that shrinks our chains below the
task evidence (Appendix~\ref{app:minimality}). \citet{lei2016rationalizing}
train a rationale selector jointly with its predictor. These works explain
or evaluate classifiers. Here sufficiency trains a selector that prunes
context for a frozen generator, which cannot co-adapt with it.

\looseness=-1
\textbf{Causal analysis of transformers.} Our interventions follow the
mediation and tracing lineage \citep{vig2020mediation, meng2022rome}.
Circuit discovery recovers minimal causal subgraphs post hoc
\citep{conmy2023acdc, bhaskar2024edgepruning}, and retrieval-head analysis
identifies heads that move or select long-context evidence
\citep{wu2024retrievalhead, wu2026expertheads}, but neither yields a per-input selector before
attention nor measures cross-run evidence consistency. Head-level stability
studies find heads differ across seeds \citep{bali2026headstability}; we
locate the stable object in the evidence sets instead. Associative recall informed the task protocol \citep{arora2023zoology,
olsson2022induction}.

\section{Limitations and conclusion}

\looseness=-1
The controlled tasks are synthetic. Counterfactual SQuAD and QASPER
are the natural-text arms, and selector generalization on QASPER stays
open. The teachers are small
(10.7M parameters) and
rotary-only (Appendix~\ref{app:protocol}). Cost is accuracy at a block
budget rather than wall clock. Router transfer covers seeds and training
length, not tokenizers or architectures. The floored estimator handles
redundancy and chains but not their combination
(Appendix~\ref{app:estimator-scope}). Labels and recovery are behavioral,
so the estimator covers evidence that is replaceable or necessary block by
block. A copyable endpoint defeats it, and counterfactual substitution
does not help.
Conflicting-fact probes cover three families; eviction adds Llama and
goes to 32B and 32K (Appendix~\ref{app:external}, generator labels;
recovery at those lengths untested). No tested router yet recovers the pretrained
multi-hop task under any label source, and on natural text the learned
selector reaches two thirds of the annotated-evidence reference on
unseen documents. The counterfactual SQuAD arm uses substitute answers and forced
choice; QASPER scores free generation. The estimator threshold is
validated on one mixture.

\looseness=-1
Sparse attention needs a target, and attention weights are the
convenient one. They fail by covering obsolete evidence and by varying
across training runs, and a selector distilled from them inherits both
omissions. Supervising on causal evidence sets repairs the routing, on
these tasks and at pretrained scale through 32B, keeping the evidence a
future question will need while accumulated attention does not.
Attention read out under a different scoring task can keep it too, so
the failures belong to particular targets, not to attention-based
compression (Appendix~\ref{app:external}).

\section*{Impact Statement}
This paper is about which blocks of a long context a model should keep. Choosing those blocks from causal evidence changes which facts survive pruning. The direct effect is on the cost and the reliability of long-context systems. The broader consequences are the ordinary ones of that line of work.

\bibliography{references}
\bibliographystyle{icml2026}

\appendix
\onecolumn
\raggedbottom
\section*{Reproducibility statement}
The code-and-results artifact submitted with this paper contains the task generators with their sufficient-set annotations, the teacher training scripts, the intervention and label code, both router implementations, and every analysis script. \texttt{reproduce.sh} reruns the stages in order. Appendix~\ref{app:repro} records the label-pipeline cost. The results files behind every number in this paper, including the appendix, are tracked in the repository as JSON, and all figures and tables are generated from those files by scripts rather than transcribed. Seed ranges for training streams, evaluation sets, and routers are fixed in code and disjoint. Each run directory stores its full configuration next to its metrics. The model components come from a small open library the repository pins as a dependency. Synthetic experiments total roughly three GPU-days on one consumer card. The pretrained experiments add roughly a GPU-day per model family using publicly available checkpoints.

\section*{Use of generative AI}
Generative AI tools were used under our direction. They implemented the task generators, the intervention and label code, both routers, and the analysis scripts, on a transformer training library we wrote. They proposed and refined hypotheses, gave feedback on the experiment design, reformatted the question-answering subset in Section~\ref{sec:pretrained} and the QASPER subset in Appendix~\ref{app:external}, and helped interpret the results. They also drafted and revised parts of the text, edited code, produced figures, suggested parameters, and searched for related work. We did not use them to develop a theoretical model, to state a mathematical claim, or to write a proof. We set the questions, the gate criteria, and the claims. Every number is regenerated from the tracked results files, and we checked the prose against those files. We are responsible for the content of this paper.

\section{Reproduction}
\label{app:repro}

\texttt{reproduce.sh} runs every stage in order, and each stage's
configuration sits next to it in that script. The label pipeline costs
two forward passes per block per example plus one per annotated set. The
KVzip comparison and the KVzip-derived labels run in their own stage
against the KVzip release at a pinned commit, in its own environment,
with the method code unchanged. The repository README lists the commit,
the library versions, and the one build flag added for newer GPUs.

\section{Task and protocol details}
\label{app:protocol}

Records are four tokens, \texttt{WRITE} family member value, padded with
filler to the block width. Queries are \texttt{QUERY} family member value
with the value supervised directly at the member token. Two protocol
findings are recorded here because both failure modes look like model
incapacity. First, placing a marker token between
the query key and the supervised value prevents retrieval circuits from
forming at any model size or learning rate we tried. The loss settles at
the guess-a-context-value level indefinitely. Second, the workable
learning rate depends on the record count. 8-record grids train at
$10^{-3}$. 16 records and up sit at the same plateau until the rate drops
to $3\times10^{-4}$. Staged curricula over record counts saturate below
half accuracy with a loss spike at each stage switch, and per-batch
mixtures of record counts never leave the plateau. Training sequences
carry eight queries. Evaluation is single-query, an exact prefix of the
training format. The aggregation answer is a count, not a parity. Parity
never trains at this scale (teachers sit at the label base rate for tens
of thousands of steps), and a control the dense model cannot solve bounds
nothing. Finally, the positional encoding is required. Teachers with
learned absolute positions never leave the same plateau at any learning
rate we tried ($10^{-4}$, $3\times10^{-4}$, $10^{-3}$; up to 45{,}000
steps), while rotary embeddings train. Every teacher in the
paper uses rotary embeddings, and the encoding axis is untested rather
than tested and passed.

\section{Hyperparameters}
\label{app:hparams}

Teachers: decoder-only, 6 layers, 8 heads, width 384, SwiGLU width 1024,
RoPE $\theta{=}10^4$, rotary sin/cos cache sized for the 4$\times$ evaluations,
AdamW with weight decay 0.01, cosine schedule from $3\times10^{-4}$ to
$3\times10^{-5}$ with 200 warmup steps, gradient clip 1.0, batch 64,
30{,}000 steps (45{,}000 for the five-task mixture), fresh examples every
step. Routers: token embedding width 64, mean-pooled block and query
summaries, a two-layer scorer, Adam at $10^{-3}$, 4{,}000 steps at batch
64 (32 on shared-GPU runs); the causal router anneals its softmin temperature
linearly to zero over the first 80\% of training. The span router for the
pretrained arm adds a cosine interaction feature between record and query
summaries and trains for 15 epochs over 2{,}000 examples. Evaluation
budgets are $\max(1, \lfloor n/10 \rfloor)$ blocks for the synthetic arm and 2
records for the pretrained arm.

\paragraph{Coverage loss.} Each example carries a family of acceptable
sets $S_1, \dots, S_m$ (one set for chains and single records, one
singleton per copy under redundancy). For router scores
$s \in \mathbb{R}^n$, let $\ell_j$ be the binary cross-entropy between
$\sigma(s)$ and the indicator of $S_j$, averaged over the $n$ blocks.
The loss is
\[
\mathcal{L} = \sum_{j} w_j \ell_j, \qquad
w = \mathrm{softmin}(\ell / \tau),
\]
with $w$ held constant in the backward pass. The temperature $\tau$ falls
linearly from 1 to 0 over the first 80\% of training, after which
$\mathcal{L} = \min_j \ell_j$. With $m{=}1$ this is ordinary multi-label
cross-entropy. The imitation loss is
$\mathrm{KL}(a \,\|\, \mathrm{softmax}(s))$, where $a$ is the teacher's
answer-position attention summed within each block and renormalized over
context blocks.

\paragraph{Backward trace.} Let $A^{(l)}$ be layer $l$'s attention,
averaged over heads, and $p$ the answer position. Starting from
$r = e_p$, for $l = L, \dots, 1$,
\[
r_j \leftarrow \max\Big(r_j,\; \max_i\, r_i A^{(l)}_{ij}\Big),
\]
which keeps the strongest single path to each position. A block's score is
$\sum_{t \in b} r_t$, and the top $k$ blocks are the label. Rollout
\citep{abnar2020quantifying} instead row-normalizes $A^{(l)} + I$ at each
layer, multiplies the layers, and reads the answer row, which sums over
paths.

\section{Minimality of behaviorally sufficient sets}
\label{app:minimality}

The greedy search (add blocks in order of masking effect until the
restricted answer is preserved, then prune) finds behaviorally sufficient
sets of 1.4 to 1.6 blocks on solved multi-hop examples, strict subsets of
the three-block chain on 86 to 100\% of them: restricted to the final
value record alone, the model often answers with the only value it can
see. The chain blocks remain individually necessary, with flip rates of
0.64 to 0.95 across seeds whose variation tracks the robustness of the
fallback rather than different routes; reading sufficient sets as the
computation's information route is sound only up to this bound.

\section{Recovery estimator ablation}
\label{app:recovered}

Table~\ref{tab:recovered} reports every estimator behind
Section~\ref{sec:recovered}. Extending the sufficiency
family with the smallest answer-preserving prefix of the masking-effect
ranking changes nothing. Behavioral answer preservation is the wrong size
criterion wherever a fallback exists (Appendix~\ref{app:minimality}).
Branching on singleton sufficiency before consulting effect sizes sends
chain examples with a strong fallback down the wrong branch, handing the
router a shortcut label. The shipped floor, 0.25 nats, is the lowest
value swept. Raising it to 4.0 nats lowers overall exact match from 0.87
to 0.75, and most of the loss is on chains (0.82 to 0.56). Floors below
0.25 nats were not tested.

The shipped estimator, in full, for one example with block set $B$ and
budget $k$:
\begin{enumerate}
\item For each $b \in B$, mask $b$ alone and record the drop
      $\Delta_b$ in the gold answer's log probability.
\item Rank $B$ by $\Delta_b$ descending; keep the ranked blocks with
      $\Delta_b \ge 0.25$ nats, at most $k$ of them.
\item If no block clears the floor, return a family of alternatives:
      each $b$ that preserves the answer when the model is restricted
      to $b$ alone is one acceptable set $\{b\}$. The coverage loss
      accepts any one of them, so the router keeps a single copy.
\item If no single block preserves the answer either, keep the block
      with the largest $\Delta_b$.
\end{enumerate}
Steps 3 and 4 are fallbacks, not the first branch. The behavioral
measurement of Section~\ref{sec:labels} branches on singleton sufficiency first; the
estimator instead prioritizes masking effects, approximating the
annotated task evidence, because running singletons first hands chain
examples with a strong fallback a shortcut label, the failure the first
paragraph describes.

% Generated by analysis/make_main_tables.py from runs/. Do not edit by hand.
\begin{table}[H]
\caption{Recovery estimators on the synthetic mixture: agreement with
the hidden annotations (8{,}000 examples) and routed accuracy of
routers trained on the recovered labels alone (three router seeds, 300
examples per condition). Cover: the recovered set contains an annotated
sufficient set. Last row: input-gradient saliency at the same budget.}
\label{tab:recovered}
\centering\footnotesize
\begin{tabular}{lccccc}
\toprule
Estimator & Match & Cover & Multi-hop & Unique & Latest \\
\midrule
Sufficiency / flip family & 0.49 & 0.59 & 0.75 & 0.97 & 0.92 \\
\quad + verified delta prefix & 0.48 & 0.60 & 0.73 & 0.96 & 0.88 \\
Top-$k$ delta, annotated size & 0.89 & 0.96 & 0.99 & 0.96 & 0.79 \\
Top-$k$ delta, budget size & 0.36 & 0.96 & 0.97 & 0.88 & 0.61 \\
Singletons, else budget delta & 0.41 & 0.62 & 0.66 & 0.97 & 0.88 \\
Floored delta, else singletons & 0.87 & 0.92 & 0.98 & 0.97 & 0.91 \\
Input-gradient saliency & 0.21 & 0.83 & 0.79 & 0.92 & 0.87 \\
\bottomrule
\end{tabular}
\end{table}

\section{Budget curves}
\label{app:ablation-budget}

\begin{figure}[H]
\begin{center}
\includegraphics[width=0.38\linewidth]{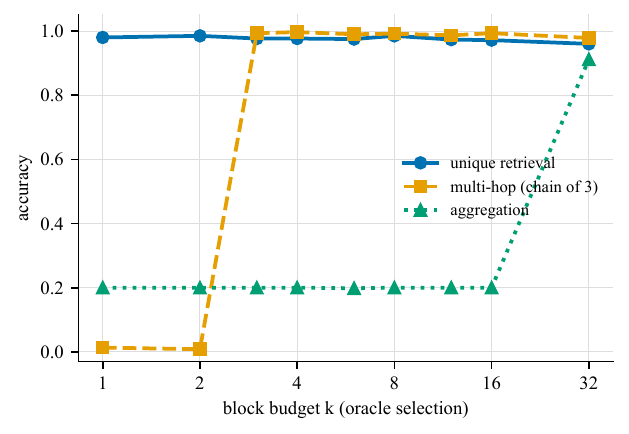}
\end{center}
\caption{Oracle accuracy against block budget $k$, annotated sufficient
blocks kept first (mean of two teachers).}
\label{fig:budget}
\end{figure}

Figure~\ref{fig:budget} gives the oracle accuracy-against-budget curve behind
the routing boundary of Section~\ref{sec:limits}, with the annotated
sufficient blocks kept first in position order, the annotated-evidence
reference; no tested router exceeds it on these tasks.

\section{Per-seed results}
\label{app:per-seed}

\begin{figure}[H]
\begin{center}
\includegraphics[width=0.38\linewidth]{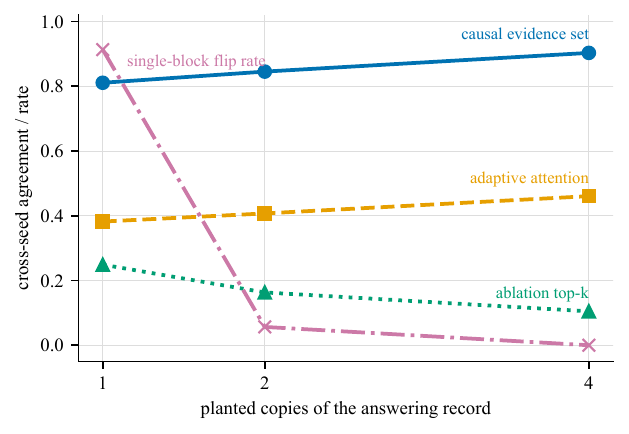}
\end{center}
\caption{Cross-seed agreement as planted copies of the answering record
increase.}
\label{fig:redundancy}
\end{figure}

\begin{figure}[H]
\begin{center}
\includegraphics[width=0.42\linewidth]{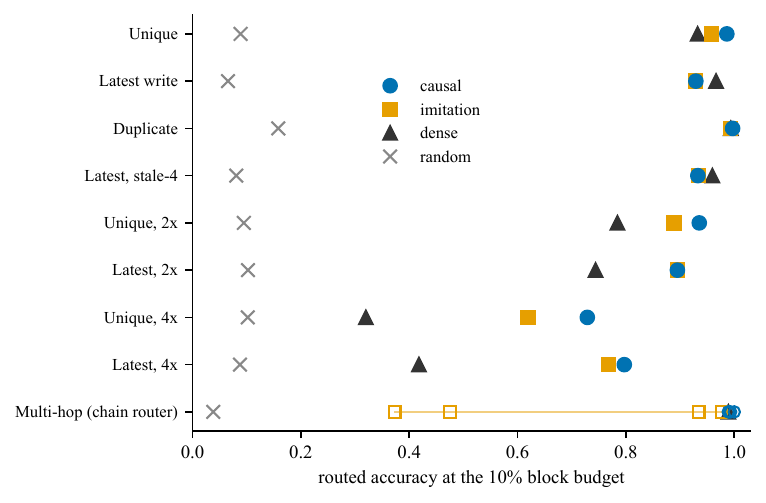}
\end{center}
\caption{Routed accuracy at the 10\% block budget. Multi-hop marks are
one per teacher, hollow, spanned by a line.}
\label{fig:routing}
\end{figure}

\begin{figure}[H]
\begin{center}
\includegraphics[width=0.38\linewidth]{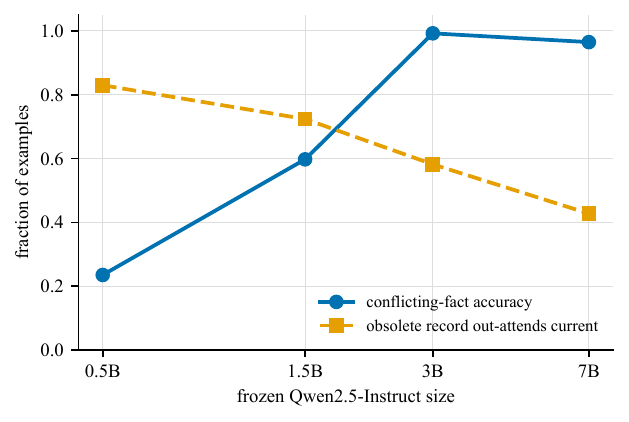}
\end{center}
\caption{Frozen Qwen2.5-Instruct on conflicting facts. Rates include the
first-record sink (Table~\ref{tab:trace-pretrained}).}
\label{fig:pretrained}
\end{figure}

Figures~\ref{fig:redundancy}--\ref{fig:pretrained} plot redundancy,
routing, and pretrained scale. Cells are in the repository's results files.

\section{Operator robustness and record-count scaling}
\label{app:overwrite}

Table~\ref{tab:overwrite} re-scores the pretrained routing selection under
two operators: the additive attention mask used throughout, and a
position-preserving content overwrite that replaces each unselected record
with a matched-length nonce record carrying a fresh name and value. In this latter case the
token count, sentence boundaries, and the query position are unchanged and
the ablated content is physically absent, not masked. Overwrite
figures average three nonce fillers, and the routers and keep-sets are held
fixed, so only the operator varies. Removing the current record flips the
answer under both operators; removing a random one does not. The
causal-over-imitation ordering and the conflicting-fact repair survive the
operator change: routing to the annotated current record lifts Gemma from
0.55 obsolete-biased dense to 1.00 under both. The trained routers narrow in
only one place: Gemma conflicting facts under overwrite. There the visible
nonce records distract a model that already answers with the obsolete value
on 45\% of these
queries dense, and a random-record overwrite flips 8\% of its answers,
against 0\% on Qwen. The mask and overwrite operators diverge on this Gemma
condition, as they do on aggregation, but the repair holds under
both. Yi-1.5-9B, unlike Gemma, answers conflicting facts
correctly dense, and it holds the causal-over-imitation ordering and near-zero
random-record flip under overwrite. The narrowing is specific to Gemma's
obsolete-answering computation, not a property of 9B scale or of the
overwrite operator. Imitation in this table keeps the first-record sink,
so the ordering reported here is against that baseline. Whether it survives the sink
correction of Appendix~\ref{app:sink} is untested under overwrite: on
conflicting facts under the mask the sink-free router already matches the
causal one, leaving no ordering to preserve. Table~\ref{tab:scaling} extends label recovery
to longer contexts: at 16, 32, 64, and 128 records the recovered labels
match annotations exactly and the recovered-label router stays at 0.96 to
1.00, at a cost of one masked forward per record plus one clean pass per
example.

% Generated by analysis/make_appendix.py from runs/. Do not edit by hand.

\begin{table}[H]
\caption{Operator robustness (150 examples; overwrite averaged over three nonce fillers). The two operators are defined in the text. Causal and Attention give the causal-supervision and attention-imitation routers' accuracy; Repair keeps only the annotated current record; Flip cur.\ and Flip rnd.\ are the rates at which removing the current or a random record changes the answer.}
\label{tab:overwrite}
\begin{center}\small
\begin{tabular}{lll ccc cc}
\toprule
Model & Task & Operator & Causal & Attention & Repair & Flip cur. & Flip rnd. \\
\midrule
Qwen2.5-3B & unique & mask & 0.99 & 0.45 & -- & 0.75 & 0.00 \\
 &  & overwrite & 1.00 & 0.47 & -- & 0.75 & 0.00 \\
 & latest & mask & 0.99 & 0.75 & 1.00 & 1.00 & 0.00 \\
 &  & overwrite & 1.00 & 0.75 & 1.00 & 1.00 & 0.00 \\
\midrule
Gemma-2-9B & unique & mask & 1.00 & 0.91 & -- & 0.80 & 0.00 \\
 &  & overwrite & 1.00 & 0.90 & -- & 0.77 & 0.00 \\
 & latest & mask & 0.99 & 0.95 & 1.00 & 1.00 & 0.04 \\
 &  & overwrite & 0.73 & 0.71 & 1.00 & 1.00 & 0.08 \\
\midrule
Yi-1.5-9B & unique & mask & 1.00 & 0.61 & -- & 0.80 & 0.00 \\
 &  & overwrite & 1.00 & 0.64 & -- & 0.80 & 0.00 \\
 & latest & mask & 0.95 & 0.83 & 1.00 & 1.00 & 0.00 \\
 &  & overwrite & 0.94 & 0.83 & 1.00 & 1.00 & 0.00 \\
\bottomrule
\end{tabular}
\end{center}
\end{table}

\begin{table}[H]
\caption{Annotation-free recovery and recovered-label routing on Qwen2.5-3B as the record count grows. Agreement is exact match of recovered labels against annotations; routing is the recovered-label router evaluated at its training record count. Recovery cost is one masked forward per record plus one clean pass per example, so the example count falls as the context grows at a fixed forward budget.}
\label{tab:scaling}
\begin{center}\small
\begin{tabular}{ccccccc}
\toprule
Records & Examples & Recovery agree & Unique route & Latest route & Forwards & Minutes \\
\midrule
16 & 2{,}000 & 1.000 & 0.99 & 0.99 & 34{,}000 & 17 \\
32 & 1{,}500 & 1.000 & 1.00 & 0.99 & 49{,}500 & 37 \\
64 & 1{,}000 & 1.000 & 0.99 & 0.98 & 65{,}000 & 43 \\
128 & 600 & 1.000 & 0.96 & 0.96 & 77{,}400 & 150 \\
\bottomrule
\end{tabular}
\end{center}
\end{table}

\section{Routing protocol: global sets against per-query selection}
\label{app:protocol-scope}

Every routing result in the main text restricts a forward pass to one block
set per example: the same blocks are visible at every query position and
every layer. Deployed block-sparse attention is weaker than that. A query
position selects its own blocks, layer by layer, so a record that the answer
position does not select may still have been read earlier in the pass. On multi-hop that difference matters, because the intermediate pointer is
resolved before the answer position is reached.

We compare four protocols at a fixed per-row budget $k$. \emph{Global} is the
main text's protocol. \emph{Per query} lets every row keep its own top-$k$
blocks, chosen from that row's attention in that layer; \emph{per query,
layer-averaged} does the same from the layer-averaged readout. \emph{Dense
prefill, sparse answer} leaves prefill unrestricted and restricts only the
answer row, which isolates the possibility that the global mask fails merely
because it deletes a record before the computation that consumed it. The
per-query mask is the global mask plus a row index: it reduces to the global
bias exactly when the selection does not vary by row, and a unit test asserts
that the two code paths then produce identical logits. Differences between the
protocols therefore come from the row dependence, not from two masks that drifted
apart.

A per-query selection is sparse in attention cost, since each row attends to
$k$ of $n$ blocks, and not sparse in context pruning, because the union over
rows and layers leaves the whole context resident. Per-query selection thins
attention. It does not prune context. Context budget counts blocks that remain usable as cross-block context. A
block our operator masks keeps its own rows, since the intervention is
position preserving, but no row outside it may read those rows at any layer,
so it supplies nothing to the rest of the pass and is not counted. Under
global selection at $k{=}3$ of 32 blocks the context budget is therefore
0.09, not 1.00.

Table~\ref{tab:per-query} gives the result. Under global selection the
attention oracle spans 0.41 to 0.98 on multi-hop across the four teachers,
reproducing the main text. Under per-query selection at the same per-row
budget it reaches 0.94 to 1.00 on every teacher, and dense prefill with a
sparse answer row reaches 0.96 to 1.00. The multi-hop failure we report is a
failure of committing to one block set before the computation runs: once
each row re-selects, the intermediate pointer is available at the position
that needs it.

Per-query selection does not prune. Its context budget is 1.00 on every
teacher, because every block stays visible to some row at some layer, so the
whole context remains resident even though each row reads a ninth of it. Causal supervision reaches 0.98 to 0.99 at a context budget of 0.09.
Attention is an adequate signal for thinning rows and an unreliable one for
choosing what to keep. Only the second is a claim about which context a
model's answer depends on. Unique retrieval separates under neither protocol (0.89
to 1.00 throughout), as in the main text: where the evidence is a single
record the teacher attends to, the two targets agree.

The per-query numbers are oracles on the teacher's own per-row
attention. They bound a learned selector; they do not show that one
trains to it. The protocols are
matched on per-row budget, not on any wall-clock or memory measure. The
translation from either budget to speed depends on kernel work we do not
attempt \citep{nawrot2025sparsefrontier}.

\section{Scope of the annotation-free estimator}
\label{app:estimator-scope}

The recovery results of Section~\ref{sec:recovered} rest on two
assumptions: that an absolute floor on the answer log-probability drop is
a meaningful cut, and that every example falls into one of the
estimator's two regimes. We test both.

\paragraph{Output calibration.} The floor is absolute, 0.25 nats, so
rescaling a model's logits changes every masking effect it is compared
against. Dividing the logits by a temperature $T$ leaves every greedy prediction
unchanged, so the model's answers, its attention, and the route its answer
depends on are identical while the log-probability drops are not.
Table~\ref{tab:temperature} sweeps $T$ over an eightfold range. The median
largest effect moves from 22.0 nats at $T{=}0.5$ to 3.5 at $T{=}4$, and the
recovered sets barely move: agreement with annotations stays between 0.845
and 0.904, and 93 to 98\% of sets are unchanged from $T{=}1$. The floor
sits one to two orders of magnitude below the effects it separates, so
rescaling those effects by four times moves few blocks across it. A model
whose evidence effects were within a few multiples of the floor would need a
normalized statistic. The floor is calibrated to this regime. It is not a constant with
meaning elsewhere.

\paragraph{Evidence topology.} The estimator's two regimes are individually
sufficient singletons and a set of individually necessary blocks. We built a case that is neither: two disjoint multi-hop chains, both entered
by the queried key, both ending at the same answer. No singleton carries a complete
task-evidence chain, because a chain needs its links; no block is necessary, because masking one
chain leaves the other. The teachers answer it dense at 0.82 to 1.00, and routing to either
annotated chain scores 0.99 to 1.00, so a correct sparse selection exists.

No estimator finds one (Table~\ref{tab:redundant}). The floored estimator's
labels contain a complete chain on 0.0 to 6.7\% of examples, the greedy
search, which the paper offers for the mixed regime, on 0.0 to 11\%, and
attention on 0.7 to 27\%. On two of the four teachers, 90 and 99\% of
examples have no block whose masking clears the floor, because the surviving
chain holds the answer in place.

The greedy failure has a second cause. Greedy returns sets that average 1.5
blocks against a three-block chain, and those sets are sufficient: the model
answers correctly from them. A chain's final record carries the answer token, and a model shown one
record and no other value reproduces it. The singleton columns of
Table~\ref{tab:redundant} separate this from noise: about one block per
example inside the chains preserves the answer alone, roughly 18\% of chain
blocks, against 1\% of blocks outside them. The estimator recovers a
behaviorally sufficient set, and under redundancy with a copyable answer
that set need not contain the route the task intends. Substituting a
counterfactual answer, as our natural-QA experiments do, does not repair
this. That substitution defeats answering from parametric memory, while the
shortcut here is copying from context, and the substituted value sits in
the terminal record exactly as the original did. Closing this case needs a
construction whose terminal record cannot be identified without traversing
the chain, so that no single record both matches the query and carries the
answer. Until then the
estimator's scope is evidence that is either replaceable block by block or
necessary block by block, with answers that cannot be produced from a single
record by copying.

\section{Budget, and which readout of attention is used (condensed in Section~\ref{sec:routers})}
\label{app:readouts}

Table~\ref{tab:budget-readouts} routes on labels at four budgets, from the
chain size to a quarter of the context.

Answer-position attention does not improve with budget. On the weakest
teacher it falls from 0.41 at $k{=}3$ to 0.26 at $k{=}8$, because the blocks
it adds are distractors rather than the middle hop, and its learned router
falls from 0.37 to 0.18 over the same range. One teacher improves, 0.51 to
0.97. Causal supervision stays at 0.98 to 1.00 at every budget on every
teacher. The gap at $k{=}3$ is not a product of the tight setting.

The chain is present in the attention weights, at rows we do not read.
Attention rollout mixes each layer with the identity and multiplies the
layers; it scores 0.24 to 0.99 at $k{=}3$, below the raw readout on two
teachers, because summing over paths lets weak paths outvote the one
carrying the answer. The backward trace keeps the strongest single
path to each position instead of summing (Appendix~\ref{app:hparams}), and scores 0.90 to 0.99 at
$k{=}3$ and 0.97 to 1.00 at $k{=}8$, close to the causal label on every
teacher. The failure is in the readout distillation uses, not in the
weights. The backward trace costs one forward pass with the attention captured, against
one clean pass plus one masked pass per block for the intervention
estimator.

The backward trace also works as a supervision target, not only as an oracle
(Table~\ref{tab:trace-supervision}). Training the chain router on the
trace's top blocks, under the coverage loss used for annotated chains and at
the settings of Table~\ref{tab:chain}, reaches 0.98 to 0.99 on every
teacher against 0.99 to 1.00 for the annotated chains, where the
answer-position target ranges from 0.36 to 0.98. On hardened chains, which
force the teacher to attend the chain, all three targets agree.

That repair does not extend to obsolete evidence. On latest write at
$k{=}3$ the trace-selected set still contains a superseded write on 0.91 to
1.00 of examples, above the raw readout's 0.70 to 0.99 and against none for
the annotated sets, and the same holds on the frozen models
(Appendix~\ref{app:sink}). Composing attention across layers recovers a
path the answer position does not expose; it does not distinguish evidence
the model has learned to discount. On these teachers, where the evidence is
a connected path, the backward trace is a sufficient target and the
interventions buy little. Where current and obsolete records compete, they
are what separates them.

\paragraph{Behavior-preserving scores.} We apply a first-order estimate
of eviction cost \citep{hwang2026bpkv} to records on the natural-language
latest-write and two-hop tasks, and compare it with exact eviction
(Table~\ref{tab:bp}). One objective is KL from the full-cache next-token
distribution. The other is the drop in gold-answer log probability.
Records leave the context-only cache before the question is computed.
Each ranking keeps its top record on latest write and its top two on the
chain. On Qwen2.5-7B, exact gold-drop leads exact KL by
\BPGoldKlLatest{} on latest write and \BPGoldKlTwohop{} on the chain.
The first-order estimate tracks exact KL, rank correlation \BPRhoKlLo{}
to \BPRhoKlHi{}. It does not track exact gold-drop scores,
\BPRhoGoldLo{} to \BPRhoGoldHi{}. On Qwen2.5-3B latest write, only the
annotated record beats random. Evicting after the question has been
prefilled raises preservation by up to \BPTimingMax{} for chain rankings
that miss the evidence. The methods reorder. The cached question states
already carry what they read. The adaptation uses a different objective,
unit, and selection from the cited method.

\section{The first-token sink in the pretrained routing comparison}
\label{app:sink}

Answer-position attention on the frozen Qwen models puts its largest record
mass on the first record of the context on every one of the 2{,}000 training
examples behind Section~\ref{sec:pretrained}, for both model sizes: the
attention sink of \citet{xiao2024streamingllm}, whose emergence
\citet{gu2025attentionsink} trace to pre-training. The same
holds for attention rollout and for the backward trace. Those readouts share
their preprocessing, so agreement alone would survive a span bug, and
Table~\ref{tab:app-span-audit} checks the preprocessing directly: every record
span decodes to its own sentence, record 0 is no longer than the median record,
and it wins under sum, mean, and max pooling alike. Prepending a sentence that
is not registered as a record moves the sink off record 0, so it sits at an
absolute position rather than on that record. The first tokens after the
sequence start absorb attention that carries no evidence. A router distilled
from that target learns to score the first record highest, and at a two-record
budget one of its two slots is spent there.

Table~\ref{tab:sink} retrains the comparison with that record removed. The
target drops the first record and renormalizes over the rest, the selection
bars it, and nothing else changes: same examples, same initialization,
optimizer, epochs, and budget. On conflicting facts the sink-free imitation
router equals the causal router at both record counts on both models, and the
fall to the random control at double length does not appear. On unique
retrieval the causal router still leads, by 10 to 30 points, and its lead
grows with record count, the pattern the synthetic length-shift results show.

Table~\ref{tab:prefix} absorbs the sink with a prepended sentence that is
not registered as a record, so every record stays selectable. Prefix 0 is
the empty string through the same code path. The empty-prefix row matches
the sink-keeping imitation column of Table~\ref{tab:budget-sink}.
Absorbing the sink leaves the causal router at 1.00 at every budget,
against the 0.96 to 0.97 the ban leaves it. The ban still recovers more of
the imitation router than the prefix does. Only unique retrieval is shown,
because the prefixes move dense accuracy on conflicting facts for the Qwen
models.

The causal no-sink column is the control. Barring the first record leaves the
causal router unchanged on conflicting facts, where the current record is never
first, and costs it three points on unique retrieval, where the answer record
is the first record on one example in sixteen. The bar is therefore not what
moves the imitation router.

The routing ordering on conflicting facts holds against an imitation
baseline that keeps the sink and does not survive its removal, so we do
not claim it. The ordering on unique retrieval under length shift
survives both. The obsolete-over-current rates of
Section~\ref{sec:pretrained} are themselves inflated by the sink whenever
an obsolete record happens to be first, which is about three examples in
sixteen by construction. Restricting to examples where no obsolete record
is first moves 3B from 0.58 to 0.48, 7B from 0.43 to 0.29, and Yi from
0.72 to 0.65 (Table~\ref{tab:trace-pretrained}). The mismatch is smaller
than the headline rate and it does not vanish.

\paragraph{Path-aware readouts here.} Appendix~\ref{app:readouts} finds a
backward trace recovering the synthetic multi-hop chain that answer-position
attention misses, and predicts it will not repair obsolete writes.
Table~\ref{tab:trace-pretrained} confirms that. The trace prefers an obsolete
record to the current one on 0.31 to 0.54 of correctly answered examples,
below the answer-position readout's 0.43 to 0.72 and far above the annotated
record, and rollout prefers one on every example of every model. Selecting the
top two records after the sink, the trace routes at 0.95 to 0.96 and
answer-position attention at 0.92 to 0.95, both near the annotated record's
1.00, which is the same conclusion the sink retraining reaches from the router
side: at this budget the mismatch costs little once the sink is gone.

% Generated by analysis/make_main_tables.py from runs/. Do not edit by hand.
\begin{table}[!htbp]
\caption{Cross-seed label agreement (mean Jaccard, ten seed pairs, 500
examples per task; labels: Section~\ref{sec:labels}). Duplicate-$m$
plants $m$ copies of the answering record.}
\label{tab:agreement}
\centering\footnotesize
\begin{tabular}{lccc}
\toprule
Task & Adaptive attention & Ablation top-$k$ & Measured evidence set \\
\midrule
Unique retrieval & 0.38 & 0.25 & \textbf{0.79} \\
Latest write & 0.43 & 0.28 & \textbf{0.81} \\
Duplicate-1 & 0.38 & 0.25 & \textbf{0.81} \\
Duplicate-2 & 0.41 & 0.16 & \textbf{0.85} \\
Duplicate-4 & 0.46 & 0.11 & \textbf{0.90} \\
\bottomrule
\end{tabular}
\end{table}

% Generated by analysis/make_main_tables.py from runs/. Do not edit by hand.
\begin{table}[!htbp]
\caption{Label type against loss form for pooling routers at the 10\%
budget, mean over teachers; bold marks the causal label under either
loss. SparseKL restricts the KL to the router's own top-$k$ set, the
primary objective of Ahmad and Yun (2026).}
\label{tab:ablation}
\centering\footnotesize
\begin{tabular}{llccc}
\toprule
Label & Loss & Multi-hop & Unique 4$\times$ & Latest 4$\times$ \\
\midrule
Attention (answer position) & KL & 0.10 & 0.64 & 0.78 \\
Attention (answer position) & SparseKL & 0.07 & 0.61 & 0.76 \\
Attention (answer position) & coverage & 0.10 & 0.66 & 0.77 \\
Attention (all positions) & KL & 0.01 & 0.09 & 0.01 \\
Causal evidence & coverage & \textbf{0.09} & \textbf{0.74} & \textbf{0.81} \\
Causal evidence & KL & \textbf{0.09} & \textbf{0.70} & \textbf{0.79} \\
\midrule
Random gates & & 0.04 & 0.10 & 0.07 \\
\bottomrule
\end{tabular}
\end{table}

\begin{figure}[H]
\begin{center}
\includegraphics[width=0.34\linewidth]{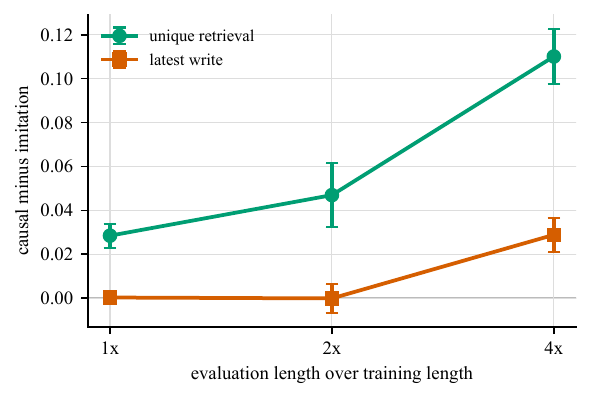}
\end{center}
\caption{Causal minus imitation (Table~\ref{tab:routing} protocol).
Error bars: one paired SE over teachers.}
\label{fig:gap}
\end{figure}

Table~\ref{tab:chain-cross} runs the label-by-loss crossing of
Table~\ref{tab:ablation} on the chain router itself, where the imitation
gap is largest. On the failing teacher, attention-derived labels route at
0.35 to 0.38 whether trained with the coverage loss or the KL loss, and
annotated labels reach 0.99 under either; on the working teacher every
arm sits at 0.97 or above. The failure follows the label, not the loss.

% Generated by analysis/make_appendix.py from runs/. Do not edit by hand.

\begin{table}[H]
\caption{Multi-hop chain routers trained on each converged teacher's own labels, three router seeds each (300 examples, $k{=}3$). Mean over router seeds with the range in brackets. Attention routing varies by more than sixty points across teachers while causal routing does not move; the main text's seed-0 comparison is the low end of that spread, not a typical case.}
\label{tab:chain-per-teacher}
\begin{center}\small
\begin{tabular}{lcc}
\toprule
Teacher & Attention (KL) & Causal (coverage) \\
\midrule
seed 0 & 0.37 [0.36--0.38] & 0.99 [0.99--0.99] \\
seed 1 & 0.98 [0.97--0.98] & 1.00 [1.00--1.00] \\
seed 2 & 0.48 [0.44--0.51] & 0.99 [0.99--0.99] \\
seed 4 & 0.93 [0.93--0.94] & 0.99 [0.99--0.99] \\
\bottomrule
\end{tabular}
\end{center}
\end{table}

\begin{table}[H]
\caption{Label-by-loss crossing on the chain router: routed accuracy
at $k{=}3$, ranges over three router seeds. The failure follows the
label: attention labels fail on the failing teacher under either
loss, and annotated labels succeed under either.}
\label{tab:chain-cross}
\begin{center}\small
\begin{tabular}{llcc}
\toprule
Teacher & Label & Coverage loss & KL loss \\
\midrule
seed 0 (imitation fails) & attention & 0.35--0.38 & 0.36--0.38 \\
 & annotated & 0.99 & 0.99 \\
\midrule
seed 1 (imitation works) & attention & 0.98--0.99 & 0.97--0.98 \\
 & annotated & 1.00 & 1.00 \\
\bottomrule
\end{tabular}
\end{center}
\end{table}

\begin{table}[H]
\caption{Multi-hop and unique-retrieval routing under per-query selection ($k{=}3$, 300 examples, ranges over the four converged teachers). Row budget is the fraction of blocks one query row may attend to. Context budget is the fraction of blocks retained as usable cross-block context, excluding the self-only islands the position-preserving mask leaves behind: a masked block keeps its own rows, but nothing outside it may read them, so it carries no context. A per-query selection is sparse per row while pruning no context.}
\label{tab:per-query}
\begin{center}\small
\begin{tabular}{lcccc}
\toprule
 & \multicolumn{2}{c}{Routed} & \multicolumn{2}{c}{Budget} \\
\cmidrule(lr){2-3}\cmidrule(lr){4-5}
Protocol & Multi-hop & Unique & Row & Context \\
\midrule
Attention, global (max heads) & 0.41--0.87 & 0.91--1.00 & 0.09 & 0.09 \\
Attention, global (mean heads) & 0.48--0.98 & 0.90--0.99 & 0.09 & 0.09 \\
Attention, per query, per layer & 0.94--1.00 & 0.91--1.00 & 0.09 & 1.00 \\
Attention, per query, layer-averaged & 0.93--0.98 & 0.91--1.00 & 0.09 & 1.00 \\
Attention, dense prefill, sparse answer & 0.96--1.00 & 0.89--0.99 & 1.00 & 1.00 \\
Causal, global & 0.98--0.99 & 0.96--0.99 & 0.09 & 0.09 \\
\bottomrule
\end{tabular}
\end{center}
\end{table}

\begin{table}[H]
\caption{Recovery under logit rescaling on the five-task teacher (seed 0, 2{,}000 examples, floor 0.25 nats). Temperature changes no greedy prediction, so the model's answers, its attention, and the route its answer depends on are identical in every row; only the log-probability drops the estimator thresholds move. Agreement is exact match against the annotated union; identical is the fraction of examples whose recovered set is unchanged from $T{=}1$.}
\label{tab:temperature}
\begin{center}\small
\begin{tabular}{cccccc}
\toprule
$T$ & Agreement & Coverage & Median max delta & Blocks over floor & Identical to $T{=}1$ \\
\midrule
0.5 & 0.845 & 0.903 & 22.0 & 1.80 & 0.980 \\
1 & 0.858 & 0.913 & 11.1 & 1.80 & -- \\
2 & 0.885 & 0.941 & 5.8 & 1.83 & 0.962 \\
4 & 0.904 & 0.965 & 3.5 & 1.87 & 0.925 \\
\bottomrule
\end{tabular}
\end{center}
\end{table}

\begin{table}[H]
\caption{Two disjoint multi-hop chains, either sufficient (300 examples, $k{=}3$; greedy on the first 100). Coverage is the fraction of examples whose recovered label contains a whole chain. Singleton columns count blocks that preserve the answer alone, inside and outside the chains.}
\label{tab:redundant}
\begin{center}\small
\begin{tabular}{lcccccc}
\toprule
 & & \multicolumn{3}{c}{Covers a chain} & \multicolumn{2}{c}{Singletons/example} \\
\cmidrule(lr){3-5}\cmidrule(lr){6-7}
Teacher & Dense & Floored & Greedy & Attention & In chain & Off chain \\
\midrule
seed 0 & 1.00 & 0.003 & 0.000 & 0.007 & 1.17 & 0.27 \\
seed 1 & 0.99 & 0.000 & 0.000 & 0.037 & 1.07 & 0.33 \\
seed 2 & 1.00 & 0.007 & 0.000 & 0.063 & 0.98 & 0.36 \\
seed 4 & 0.82 & 0.067 & 0.110 & 0.270 & 1.19 & 0.29 \\
\bottomrule
\end{tabular}
\end{center}
\end{table}

\begin{table}[H]
\caption{Multi-hop routing against block budget, ranges over the four converged teachers (300 examples, 32 blocks). Every row routes on labels directly.}
\label{tab:budget-readouts}
\begin{center}\small
\begin{tabular}{lcccc}
\toprule
Selection & $k{=}3$ & $k{=}4$ & $k{=}5$ & $k{=}8$ \\
 & (0.09) & (0.12) & (0.16) & (0.25) \\
\midrule
Attention, answer position (max heads) & 0.41--0.87 & 0.30--0.96 & 0.24--0.98 & 0.26--1.00 \\
Attention, answer position (mean heads) & 0.48--0.98 & 0.34--0.98 & 0.29--0.99 & 0.28--1.00 \\
Attention rollout & 0.24--0.99 & 0.31--0.99 & 0.57--0.99 & 0.94--1.00 \\
Attention, backward trace & 0.90--0.99 & 0.96--0.99 & 0.95--1.00 & 0.97--1.00 \\
Causal, ablation ranking & 0.98--0.99 & 0.98--1.00 & 0.98--1.00 & 0.98--1.00 \\
Annotated chain, padded & 0.99--1.00 & 0.98--1.00 & 0.98--1.00 & 0.99--1.00 \\
Random & 0.04--0.08 & 0.03--0.06 & 0.05--0.07 & 0.05--0.08 \\
\bottomrule
\end{tabular}
\end{center}
\end{table}

\begin{table}[H]
\caption{Each arm trained on the teacher it is evaluated on, versus the published seed-0 protocol of Table~\ref{tab:routing}. The seed-0 columns are an independent replication, not the same checkpoints. $\Delta$ is causal minus imitation paired within each teacher, $\pm$ one standard error over the five seeds.}
\label{tab:app-own-teacher}
\begin{center}\footnotesize\setlength{\tabcolsep}{3pt}
\begin{tabular}{lcccccc}
\toprule
 & \multicolumn{3}{c}{Trained on seed 0} & \multicolumn{3}{c}{Trained on own teacher} \\
\cmidrule(lr){2-4}\cmidrule(lr){5-7}
Condition & Imitation & Causal & $\Delta$ & Imitation & Causal & $\Delta$ \\
\midrule
Unique retrieval & 0.951 & 0.982 & $+0.031 \pm 0.006$ & 0.930 & 0.981 & $+0.051 \pm 0.012$ \\
Latest write & 0.929 & 0.929 & $+0.000 \pm 0.001$ & 0.925 & 0.929 & $+0.004 \pm 0.003$ \\
Duplicate evidence & 0.992 & 0.996 & $+0.004 \pm 0.003$ & 0.987 & 0.996 & $+0.009 \pm 0.002$ \\
Latest write, stale-4 & 0.934 & 0.934 & $+0.000 \pm 0.002$ & 0.926 & 0.930 & $+0.004 \pm 0.005$ \\
Unique, family distractors & 0.961 & 0.986 & $+0.025 \pm 0.009$ & 0.938 & 0.985 & $+0.046 \pm 0.015$ \\
Unique, 2$\times$ length & 0.870 & 0.928 & $+0.058 \pm 0.011$ & 0.841 & 0.917 & $+0.076 \pm 0.018$ \\
Latest, 2$\times$ length & 0.886 & 0.890 & $+0.004 \pm 0.005$ & 0.865 & 0.891 & $+0.026 \pm 0.012$ \\
Unique, 4$\times$ length & 0.615 & 0.722 & $+0.107 \pm 0.015$ & 0.570 & 0.716 & $+0.146 \pm 0.019$ \\
Latest, 4$\times$ length & 0.748 & 0.796 & $+0.048 \pm 0.006$ & 0.726 & 0.790 & $+0.064 \pm 0.011$ \\
\bottomrule
\end{tabular}
\end{center}
\end{table}

\begin{table}[H]
\caption{Tuning guard: the imitation router retrained at other learning rates and at triple steps (means over the three-seed teachers of that phase).}
\label{tab:app-guard}
\begin{center}\small
\begin{tabular}{llccc}
\toprule
Retuning & Condition & Imitation & Causal & Gap \\
\midrule
lr 3e-4 & unique 4x & 0.64 & 0.71 & +0.08 \\
lr 3e-4 & latest 4x & 0.78 & 0.79 & +0.00 \\
lr 3e-3 & unique 4x & 0.61 & 0.74 & +0.13 \\
lr 3e-3 & latest 4x & 0.76 & 0.78 & +0.03 \\
3x steps & unique 4x & 0.61 & 0.69 & +0.08 \\
3x steps & latest 4x & 0.77 & 0.77 & +0.00 \\
\bottomrule
\end{tabular}
\end{center}
\end{table}

\begin{table}[H]
\caption{Every router against every teacher (3 router seeds, 5 teachers, 10\% budget). Source is the spread of a destination's score across the teachers whose labels supervised it. Seed is the same statistic across initializations. Off is the mean over source-teacher pairs that differ, against the diagonal where a router is deployed on the teacher that supervised it.}
\label{tab:app-transfer}
\begin{center}\footnotesize
\begin{tabular}{lcccccccc}
\toprule
 & \multicolumn{4}{c}{Imitation} & \multicolumn{4}{c}{Causal} \\
\cmidrule(lr){2-5}\cmidrule(lr){6-9}
Condition & diag & off & source & seed & diag & off & source & seed \\
\midrule
Unique retrieval & 0.929 & 0.951 & 0.0111 & 0.0024 & 0.985 & 0.985 & 0.0011 & 0.0047 \\
Latest write & 0.925 & 0.929 & 0.0020 & 0.0004 & 0.929 & 0.929 & 0.0007 & 0.0009 \\
Duplicate evidence & 0.987 & 0.992 & 0.0028 & 0.0012 & 0.997 & 0.997 & 0.0004 & 0.0011 \\
Latest write, stale-4 & 0.930 & 0.931 & 0.0027 & 0.0011 & 0.934 & 0.933 & 0.0012 & 0.0014 \\
Unique, family distractors & 0.939 & 0.959 & 0.0104 & 0.0017 & 0.988 & 0.988 & 0.0010 & 0.0035 \\
Unique, 2$\times$ length & 0.844 & 0.869 & 0.0178 & 0.0047 & 0.933 & 0.934 & 0.0024 & 0.0094 \\
Latest, 2$\times$ length & 0.866 & 0.879 & 0.0120 & 0.0033 & 0.899 & 0.896 & 0.0024 & 0.0070 \\
Unique, 4$\times$ length & 0.565 & 0.592 & 0.0233 & 0.0069 & 0.728 & 0.727 & 0.0053 & 0.0121 \\
Latest, 4$\times$ length & 0.731 & 0.740 & 0.0228 & 0.0052 & 0.793 & 0.797 & 0.0041 & 0.0100 \\
\bottomrule
\end{tabular}
\end{center}
\end{table}

\begin{table}[H]
\caption{Pretrained record routing at a two-record budget (150 examples per cell), with the first-token sink kept and removed. No-sink drops the first record from the target, renormalizes the rest, and bars it from selection.}
\label{tab:sink}
\begin{center}\small
\begin{tabular}{llcccccc}
\toprule
 & & & \multicolumn{2}{c}{Imitation} & \multicolumn{2}{c}{Causal} & \\
\cmidrule(lr){4-5}\cmidrule(lr){6-7}
Model & Condition & Dense & published & no sink & published & no sink & Random \\
\midrule
Qwen2.5-3B & unique, 16 & 1.00 & 0.45 & \textbf{0.79} & 0.99 & 0.96 & 0.40 \\
 & unique, 32 & 1.00 & 0.31 & \textbf{0.71} & 1.00 & 0.97 & 0.29 \\
 & conflicting, 16 & 0.99 & 0.75 & \textbf{0.99} & 0.99 & 0.99 & 0.32 \\
 & conflicting, 32 & 0.99 & 0.25 & \textbf{0.97} & 0.97 & 0.97 & 0.29 \\
\midrule
Qwen2.5-7B & unique, 16 & 1.00 & 0.69 & \textbf{0.89} & 0.99 & 0.95 & 0.36 \\
 & unique, 32 & 1.00 & 0.27 & \textbf{0.87} & 1.00 & 0.99 & 0.35 \\
 & conflicting, 16 & 0.98 & 0.87 & \textbf{0.97} & 0.97 & 0.97 & 0.25 \\
 & conflicting, 32 & 0.96 & 0.23 & \textbf{0.93} & 0.93 & 0.93 & 0.24 \\
\bottomrule
\end{tabular}
\end{center}
\end{table}

\begin{table}[H]
\caption{Readouts on frozen models, conflicting facts, on the examples each model answers correctly, over the same 400 examples as Table~\ref{tab:pretrained}. Obsolete$>$current is the rate that some obsolete record outscores the current one, first over all of these examples and then over those where no obsolete record occupies the first-record sink. Routed keeps the top two records; no-sink keeps the top two after the first record. The annotated record scores 1.00 everywhere.}
\label{tab:trace-pretrained}
\begin{center}\small
\begin{tabular}{llcccc}
\toprule
 & & \multicolumn{2}{c}{Obsolete $>$ current} & \multicolumn{2}{c}{Routed} \\
\cmidrule(lr){3-4}\cmidrule(lr){5-6}
Model & Readout & all & no sink & top 2 & no sink \\
\midrule
Qwen2.5-3B & attention & 0.58 & 0.48 & 0.50 & 0.95 \\
 & trace & 0.47 & 0.34 & 0.63 & 0.95 \\
 & rollout & 1.00 & 1.00 & 0.17 & 0.15 \\
\midrule
Qwen2.5-7B & attention & 0.43 & 0.29 & 0.70 & 0.95 \\
 & trace & 0.31 & 0.15 & 0.84 & 0.95 \\
 & rollout & 1.00 & 1.00 & 0.17 & 0.14 \\
\midrule
Yi-1.5-9B & attention & 0.72 & 0.65 & 0.37 & 0.92 \\
 & trace & 0.54 & 0.43 & 0.58 & 0.96 \\
 & rollout & 1.00 & 1.00 & 0.16 & 0.17 \\
\bottomrule
\end{tabular}
\end{center}
\end{table}

\begin{table}[H]
\caption{Pretrained record routing against the budget, with the first-token sink kept and removed (150 examples per cell, 16 records). No-sink drops record 0 from the target and bars it from selection. Recall is the sink-keeping imitation router's rate of holding the current record.}
\label{tab:budget-sink}
\begin{center}\small
\begin{tabular}{llccccccc}
\toprule
 & & & \multicolumn{2}{c}{Imitation} & \multicolumn{2}{c}{Causal} & & \\
\cmidrule(lr){4-5}\cmidrule(lr){6-7}
Model & Condition & Dense & kept & no sink & kept & no sink & Random & Recall \\
\midrule
Qwen2.5-3B & unique, $k{=}1$ & 1.00 & 0.29 & 0.53 & 0.99 & 0.97 & -- & -- \\
 & unique, $k{=}2$ & 1.00 & 0.45 & 0.79 & 0.99 & 0.96 & -- & -- \\
 & unique, $k{=}4$ & 1.00 & 0.90 & 0.93 & 1.00 & 0.97 & -- & -- \\
 & conflicting, $k{=}1$ & 0.99 & 0.19 & 0.75 & 0.91 & 0.91 & 0.24 & 0.00 \\
 & conflicting, $k{=}2$ & 0.99 & 0.75 & 0.99 & 0.99 & 0.99 & 0.33 & 0.75 \\
 & conflicting, $k{=}4$ & 0.99 & 0.99 & 0.97 & 0.96 & 0.95 & 0.39 & 1.00 \\
\midrule
Qwen2.5-7B & unique, $k{=}1$ & 1.00 & 0.29 & 0.76 & 0.99 & 0.97 & -- & -- \\
 & unique, $k{=}2$ & 1.00 & 0.69 & 0.89 & 0.99 & 0.95 & -- & -- \\
 & unique, $k{=}4$ & 1.00 & 0.99 & 0.95 & 1.00 & 0.97 & -- & -- \\
 & conflicting, $k{=}1$ & 0.98 & 0.19 & 0.93 & 0.91 & 0.91 & 0.21 & 0.00 \\
 & conflicting, $k{=}2$ & 0.98 & 0.87 & 0.97 & 0.97 & 0.97 & 0.31 & 0.87 \\
 & conflicting, $k{=}4$ & 0.98 & 0.97 & 0.97 & 0.95 & 0.95 & 0.36 & 1.00 \\
\midrule
Yi-1.5-9B & unique, $k{=}1$ & 1.00 & 0.25 & 0.65 & 1.00 & 0.96 & -- & -- \\
 & unique, $k{=}2$ & 1.00 & 0.61 & 0.94 & 1.00 & 0.96 & -- & -- \\
 & unique, $k{=}4$ & 1.00 & 0.99 & 0.97 & 1.00 & 0.96 & -- & -- \\
 & conflicting, $k{=}1$ & 1.00 & 0.18 & 0.85 & 0.87 & 0.87 & 0.23 & 0.00 \\
 & conflicting, $k{=}2$ & 1.00 & 0.83 & 0.95 & 0.95 & 0.95 & 0.25 & 0.83 \\
 & conflicting, $k{=}4$ & 1.00 & 0.95 & 0.93 & 0.99 & 0.95 & 0.41 & 1.00 \\
\bottomrule
\end{tabular}
\end{center}
\end{table}

\begin{table}[H]
\caption{Correcting the sink by absorption rather than prohibition. A sentence is prepended without being registered as a record. Prefix 0 is the empty string. The sink column is the rate at which record 0 holds the largest record mass over 600 training examples. Training and evaluation match Table~\ref{tab:budget-sink}. Only unique retrieval is shown.}
\label{tab:prefix}
\begin{center}\footnotesize
\begin{tabular}{llccccccc}
\toprule
 & & & \multicolumn{3}{c}{Unique, imitation} & \multicolumn{3}{c}{Unique, causal} \\
\cmidrule(lr){4-6}\cmidrule(lr){7-9}
Model & Prefix & Sink & $k{=}1$ & $k{=}2$ & $k{=}4$ & $k{=}1$ & $k{=}2$ & $k{=}4$ \\
\midrule
Qwen2.5-3B & none & 1.00 & 0.29 & 0.45 & 0.90 & 0.99 & 0.99 & 1.00 \\
 & 1 & 0.36 & 0.45 & 0.89 & 0.98 & 1.00 & 1.00 & 1.00 \\
 & 2 & 0.33 & 0.39 & 0.70 & 0.93 & 1.00 & 1.00 & 1.00 \\
 & 3 & 0.39 & 0.33 & 0.43 & 0.81 & 1.00 & 1.00 & 1.00 \\
\midrule
Qwen2.5-7B & none & 1.00 & 0.29 & 0.69 & 0.99 & 0.99 & 0.99 & 1.00 \\
 & 1 & 0.51 & 0.33 & 0.55 & 0.83 & 1.00 & 1.00 & 1.00 \\
 & 2 & 0.44 & 0.37 & 0.54 & 0.91 & 1.00 & 1.00 & 1.00 \\
 & 3 & 0.39 & 0.43 & 0.56 & 0.86 & 1.00 & 1.00 & 1.00 \\
\midrule
Yi-1.5-9B & none & 1.00 & 0.25 & 0.61 & 0.99 & 1.00 & 1.00 & 1.00 \\
 & 1 & 0.42 & 0.31 & 0.77 & 0.99 & 1.00 & 1.00 & 1.00 \\
 & 2 & 0.07 & 0.55 & 0.81 & 0.99 & 1.00 & 1.00 & 1.00 \\
 & 3 & 0.29 & 0.39 & 0.65 & 0.99 & 1.00 & 1.00 & 1.00 \\
\bottomrule
\end{tabular}
\end{center}
\end{table}

\begin{table}[H]
\caption{The backward trace as a supervision target rather than an oracle. Each row trains the chain router on the labels named, at the settings of Table~\ref{tab:chain}. The last column is the rate at which the selected set still contains a superseded write on latest write at $k{=}3$.}
\label{tab:trace-supervision}
\begin{center}\small
\begin{tabular}{lccc}
\toprule
Supervision & Chains & Hardened chains & Keeps an obsolete write \\
\midrule
Annotated evidence & 0.99--1.00 & 0.99--1.00 & 0.00 \\
Backward trace & 0.98--0.99 & 0.98--0.99 & 0.91--1.00 \\
Answer-position attention & 0.36--0.98 & 0.97 & 0.70--0.99 \\
\bottomrule
\end{tabular}
\end{center}
\end{table}

\begin{table}[H]
\caption{Readouts as selection oracles on the pretrained multi-hop task (200 examples, 16 records; rates on the examples each model solves dense). Each row keeps a readout's top two records. Removing the pointer record leaves 0.21, 0.19, 0.40 against random-pair floors of 0.23, 0.27, 0.28, so both records are necessary for the Qwen models while Yi keeps a small shortcut. Recall columns are the pointer, the value, and both.}
\label{tab:two-hop-readouts}
\begin{center}\small
\begin{tabular}{llcccc}
\toprule
Model & Readout & Routed & Pointer & Value & Both \\
\midrule
Qwen2.5-7B (190/200 solved) & answer-position attention & 0.34 & 0.43 & 0.25 & 0.05 \\
 & rollout & 0.21 & 0.13 & 0.11 & 0.02 \\
 & trace & 0.34 & 0.60 & 0.26 & 0.07 \\
 & annotated & 1.00 & 1.00 & 1.00 & 1.00 \\
 & random & 0.22 & 0.16 & 0.12 & 0.01 \\
\midrule
Qwen2.5-3B (161/200 solved) & answer-position attention & 0.34 & 0.17 & 0.17 & 0.01 \\
 & rollout & 0.25 & 0.13 & 0.11 & 0.02 \\
 & trace & 0.33 & 0.17 & 0.14 & 0.01 \\
 & annotated & 1.00 & 1.00 & 1.00 & 1.00 \\
 & random & 0.24 & 0.11 & 0.12 & 0.01 \\
\midrule
Yi-1.5-9B (179/200 solved) & answer-position attention & 0.48 & 0.21 & 0.53 & 0.06 \\
 & rollout & 0.21 & 0.12 & 0.10 & 0.02 \\
 & trace & 0.67 & 0.22 & 0.78 & 0.10 \\
 & annotated & 1.00 & 1.00 & 1.00 & 1.00 \\
 & random & 0.27 & 0.14 & 0.11 & 0.01 \\
\bottomrule
\end{tabular}
\end{center}
\end{table}

\begin{table}[H]
\caption{Selected heads as selection oracles on the pretrained multi-hop task, on the test examples of Table~\ref{tab:two-hop-readouts} at the chain budget. Heads are selected on 200 held-out examples, each shown in 4 record orders: a head hits when its most-attended record at the answer position is a chain record, and heads are ranked by the share of examples on which they hit in every order, then by hit rate. Top $m$ pools the $m$ best heads; roles keeps the top record of the best pointer head and of the best value head. Best head is chosen on the test examples themselves and bounds any selection. Rows marked interventions select heads against the records whose masking alone flips the answer instead of the annotated chain; that set equals the annotated chain on 39--58\% of selection examples. The top head is the same under both selections except for Qwen2.5-14B, whose two best heads swap order.}
\label{tab:expert-heads}
\begin{center}\small
\begin{tabular}{llcccc}
\toprule
Model & Readout & Routed & Pointer & Value & Both \\
\midrule
Qwen2.5-7B (190/200 solved) & pooled attention & 0.34 & 0.44 & 0.25 & 0.06 \\
 & top head & 0.72 & 0.94 & 0.68 & 0.65 \\
 & top 4 heads & 0.75 & 0.93 & 0.73 & 0.68 \\
 & top 16 heads & 0.56 & 0.88 & 0.49 & 0.41 \\
 & roles & 0.76 & 0.89 & 0.74 & 0.66 \\
 & annotated & 1.00 & 1.00 & 1.00 & 1.00 \\
 & top head, interventions & 0.72 & 0.94 & 0.68 & 0.65 \\
 & top 4, interventions & 0.75 & 0.93 & 0.73 & 0.68 \\
 & top 16, interventions & 0.56 & 0.88 & 0.49 & 0.41 \\
 & best head (post hoc) & & & & 0.67 \\
\midrule
Qwen2.5-3B (159/200 solved) & pooled attention & 0.34 & 0.19 & 0.14 & 0.01 \\
 & top head & 0.76 & 0.52 & 0.81 & 0.39 \\
 & top 4 heads & 0.74 & 0.33 & 0.88 & 0.30 \\
 & top 16 heads & 0.65 & 0.23 & 0.82 & 0.18 \\
 & roles & 0.76 & 0.47 & 0.79 & 0.36 \\
 & annotated & 1.00 & 1.00 & 1.00 & 1.00 \\
 & top head, interventions & 0.76 & 0.52 & 0.81 & 0.39 \\
 & top 4, interventions & 0.74 & 0.33 & 0.88 & 0.30 \\
 & top 16, interventions & 0.65 & 0.23 & 0.82 & 0.18 \\
 & best head (post hoc) & & & & 0.39 \\
\midrule
Yi-1.5-9B (179/200 solved) & pooled attention & 0.48 & 0.21 & 0.53 & 0.06 \\
 & top head & 0.87 & 0.80 & 0.88 & 0.71 \\
 & top 4 heads & 0.86 & 0.89 & 0.85 & 0.75 \\
 & top 16 heads & 0.82 & 0.77 & 0.84 & 0.63 \\
 & roles & 0.73 & 0.67 & 0.78 & 0.49 \\
 & annotated & 1.00 & 1.00 & 1.00 & 1.00 \\
 & top head, interventions & 0.87 & 0.80 & 0.88 & 0.71 \\
 & top 4, interventions & 0.86 & 0.75 & 0.88 & 0.65 \\
 & top 16, interventions & 0.82 & 0.72 & 0.85 & 0.59 \\
 & best head (post hoc) & & & & 0.71 \\
\midrule
Qwen2.5-14B (200/200 solved) & pooled attention & 0.68 & 0.12 & 0.82 & 0.07 \\
 & top head & 0.78 & 0.57 & 0.89 & 0.47 \\
 & top 4 heads & 0.78 & 0.21 & 0.97 & 0.21 \\
 & top 16 heads & 0.83 & 0.38 & 0.96 & 0.35 \\
 & roles & 0.84 & 0.50 & 0.94 & 0.45 \\
 & annotated & 1.00 & 1.00 & 1.00 & 1.00 \\
 & top head, interventions & 0.76 & 0.13 & 0.97 & 0.13 \\
 & top 4, interventions & 0.78 & 0.21 & 0.97 & 0.21 \\
 & top 16, interventions & 0.81 & 0.31 & 0.97 & 0.29 \\
 & best head (post hoc) & & & & 0.56 \\
\bottomrule
\end{tabular}
\end{center}
\end{table}

\begin{table}[H]
\caption{Record rankings on the natural-language latest-write and two-hop tasks, scored by exact eviction or by the first-order estimate adapted from \citet{hwang2026bpkv}, under next-token KL or the gold-answer drop. Each ranking keeps its top one (latest write) or two (chain) records; the cells are the share of solved examples whose forced choice survives evicting the rest. Pre evicts from the context-only cache before the question is computed; post evicts after the question has been prefilled. Three disjoint sets of 100 examples.}
\label{tab:bp}
\begin{center}\small
\begin{tabular}{llcc}
\toprule
Model and task & Ranking & pre & post \\
\midrule
Qwen2.5-3B, latest write (294) & exact, gold & 0.21 & 0.21 \\
 & exact, KL & 0.24 & 0.23 \\
 & first-order, gold & 0.07 & 0.07 \\
 & first-order, KL & 0.22 & 0.20 \\
 & pooled attention & 0.21 & 0.20 \\
 & annotated & 1.00 & 1.00 \\
 & random & 0.31 & 0.28 \\
\midrule
Qwen2.5-7B, latest write (300) & exact, gold & 0.67 & 0.59 \\
 & exact, KL & 0.25 & 0.33 \\
 & first-order, gold & 0.22 & 0.20 \\
 & first-order, KL & 0.25 & 0.22 \\
 & pooled attention & 0.22 & 0.20 \\
 & annotated & 1.00 & 1.00 \\
 & random & 0.28 & 0.29 \\
\midrule
Qwen2.5-3B, two-hop (248) & exact, gold & 0.34 & 0.66 \\
 & exact, KL & 0.29 & 0.34 \\
 & first-order, gold & 0.60 & 0.73 \\
 & first-order, KL & 0.31 & 0.35 \\
 & pooled attention & 0.31 & 0.33 \\
 & annotated & 1.00 & 1.00 \\
 & random & 0.22 & 0.24 \\
\midrule
Qwen2.5-7B, two-hop (296) & exact, gold & 0.67 & 0.78 \\
 & exact, KL & 0.32 & 0.61 \\
 & first-order, gold & 0.60 & 0.76 \\
 & first-order, KL & 0.41 & 0.44 \\
 & pooled attention & 0.70 & 0.81 \\
 & annotated & 0.99 & 1.00 \\
 & random & 0.29 & 0.29 \\
\bottomrule
\end{tabular}
\end{center}
\end{table}

\begin{table}[H]
\caption{Learned routers on the pretrained multi-hop task at the chain budget ($k{=}2$, 150 examples). Chain recall is the rate of keeping both chain records. The mixer variant runs record summaries and the query through a two-layer transformer. Random chain recall is $1/\binom{16}{2} \approx 0.01$.}
\label{tab:two-hop-router}
\begin{center}\small
\begin{tabular}{llcc}
\toprule
Model & Router / label & Routed & Chain recall \\
\midrule
Qwen2.5-7B & span, annotated & 0.21 & 0.07 \\
 & span, trace & 0.21 & 0.03 \\
 & span, attention & 0.21 & 0.02 \\
 & mixer, annotated & 0.23 & 0.08 \\
 & mixer, annotated (20k) & 0.22 & 0.10 \\
 & mixer, trace & 0.21 & 0.03 \\
 & mixer, attention & 0.23 & 0.02 \\
 & annotated pair (oracle) & 1.00 & 1.00 \\
 & random & 0.23 & 0.01 \\
\midrule
Qwen2.5-3B & span, annotated & 0.21 & 0.07 \\
 & span, trace & 0.27 & 0.00 \\
 & span, attention & 0.25 & 0.03 \\
 & annotated pair (oracle) & 1.00 & 1.00 \\
 & random & 0.22 & 0.01 \\
\midrule
Yi-1.5-9B & span, annotated & 0.28 & 0.13 \\
 & span, trace & 0.23 & 0.03 \\
 & span, attention & 0.20 & 0.03 \\
 & annotated pair (oracle) & 1.00 & 1.00 \\
 & random & 0.23 & 0.01 \\
\bottomrule
\end{tabular}
\end{center}
\end{table}

\begin{table}[H]
\caption{Training-free selection statistics as oracles on the pretrained tasks (150 examples per cell, 16 records, matching Table~\ref{tab:budget-sink}). Accumulated sums attention into each record over every query position and layer. Window sums it from the question tokens only.}
\label{tab:eviction}
\begin{center}\small
\begin{tabular}{llcccc}
\toprule
Model & Condition & Accumulated & Window & Random \\
\midrule
Qwen2.5-3B & unique, $k{=}1$ & 0.29 & 0.29 & 0.31 \\
 & unique, $k{=}2$ & 0.31 & 0.39 & 0.34 \\
 & unique, $k{=}4$ & 0.41 & 0.66 & 0.43 \\
 & conflicting, $k{=}1$ & 0.19 & 0.19 & 0.27 \\
 & conflicting, $k{=}2$ & 0.17 & 0.27 & 0.25 \\
 & conflicting, $k{=}4$ & 0.05 & 0.85 & 0.33 \\
\midrule
Qwen2.5-7B & unique, $k{=}1$ & 0.29 & 0.29 & 0.31 \\
 & unique, $k{=}2$ & 0.35 & 0.45 & 0.32 \\
 & unique, $k{=}4$ & 0.44 & 0.87 & 0.41 \\
 & conflicting, $k{=}1$ & 0.19 & 0.19 & 0.25 \\
 & conflicting, $k{=}2$ & 0.17 & 0.51 & 0.27 \\
 & conflicting, $k{=}4$ & 0.08 & 0.97 & 0.39 \\
\midrule
Yi-1.5-9B & unique, $k{=}1$ & 0.25 & 0.25 & 0.34 \\
 & unique, $k{=}2$ & 0.28 & 0.40 & 0.31 \\
 & unique, $k{=}4$ & 0.43 & 0.87 & 0.45 \\
 & conflicting, $k{=}1$ & 0.18 & 0.18 & 0.25 \\
 & conflicting, $k{=}2$ & 0.15 & 0.45 & 0.27 \\
 & conflicting, $k{=}4$ & 0.05 & 0.75 & 0.33 \\
\bottomrule
\end{tabular}
\end{center}
\end{table}

\begin{table}[H]
\caption{Audit of the record spans behind every pretrained readout. Each row checks that every record span decodes to its sentence, that spans are ordered and non-overlapping, and that record 0 starts at token 0. The last columns prepend a neutral sentence that is not registered as a record.}
\label{tab:app-span-audit}
\begin{center}\small
\begin{tabular}{lccccc}
\toprule
 & & Record 0 tokens & \multicolumn{2}{c}{Record 0 is argmax} & Records above \\
\cmidrule(lr){4-5}
Model & Integrity & (median record) & plain & prefixed & the prefix \\
\midrule
Qwen2.5-3B & pass & 12.6 (12.9) & 10/10 & 0/10 & 0 \\
Qwen2.5-7B & pass & 12.6 (12.9) & 10/10 & 0--1/10 & 0 \\
Yi-1.5-9B & pass & 13.0 (13.0) & 10/10 & 0/10 & 0 \\
\bottomrule
\end{tabular}
\end{center}
\end{table}

\section{Block eviction at pretrained scale}
\label{app:external}

This appendix tests the paper's supervision claim on frozen pretrained
models at benchmark scale. The selector is a block router over long
contexts, trained once on attention labels and once on causal labels,
with everything else held fixed. Three tasks come from the RULER
generators \citep{hsieh2024ruler}: single-needle retrieval, multikey
retrieval, and variable tracking, which chains five assignments. A fourth
task is a register log in which one register is written twice and only
the later write is correct. Contexts are cut into blocks of 128 tokens.
A policy retains a fixed fraction of blocks, the question remains
visible, and an example counts as preserved when a short greedy
continuation, decoded with no access to the reference answer, still
contains every reference answer string. Teacher-forced token
match is also recorded in the results files; it undercounts models that
phrase the answer in different tokens, so every table here reports the
generative rule. Routers train
at 4K context. The primary scoring model is Qwen2.5-7B-Instruct,
evaluated at 8K, 16K, and 32K. Qwen2.5-3B-Instruct repeats 8K and 16K.
Llama-3.1-8B-Instruct and Qwen2.5-32B-Instruct repeat the pre-query
protocol at 8K and 16K.

The router pools each block from the model's input embeddings and an
early hidden layer, encodes the pooled summaries with a two-layer
transformer, and scores each block. Imitation distills the
model's accumulated attention over blocks. Causal supervision learns binary
labels marking the blocks that generator ground truth identifies as
evidence. Both routers share features, data, ordering, initialization, and
optimizer. The baselines are the accumulated-attention policy of H2O
\citep{zhang2023h2o}, the observation-window policy of SnapKV
\citep{li2024snapkv}, a policy using the selection statistic of MomentKV
\citep{li2026momentkv}, recency, and random blocks, all acting once at
prefill and at block granularity. Tables and figures mark these
adaptations H2O-style, SnapKV-style, and MomentKV-style. None is the
native system.

\paragraph{Query-aware eviction.} When every policy sees the question,
SnapKV is close to the generator evidence set on retrieval and on the
register log (Table~\ref{tab:qa}). The question names its target on these tasks, and
the window statistic finds it. The causal router leads only on variable
tracking at the 10\% budget, where the evidence is a chain the question
does not name. The imitation router fails on every condition
despite distilling its target closely. Imitation is the weak
link, not the architecture.

\paragraph{Pre-query eviction.} Deployed caches often shrink before the
question exists. In this regime the router receives no query features,
its labels mark every block a future question could need, and the
attention statistics are computed from context tokens alone. SnapKV
cannot run here and is shown only as a reference. An oracle keeps the
query-free evidence blocks identified by the generator; when the set
exceeds the budget it is truncated in document order, so it is an upper
bound only at budgets that hold the full set. Where the causal router
beats that oracle, it has ranked the evidence blocks ahead of document
order, which often drops the ones that matter.
Table~\ref{tab:pq16k} is the 16K grid. The 8K grid has the same
ordering. That grid, and the repeated model and seed grids, are in the
repository's results files. On
multikey retrieval at the 10\% budget the causal router preserves 0.95
of dense-correct answers against 0.10 for H2O at 8K. On the register
log, H2O retains the correction on 22\% of examples at 8K, and
evicting it is fatal: preservation is 1.00 when the queried evidence
is kept and 0.00 when it is dropped. The causal router retains the
correction on 96\% or more.
Accumulated attention concentrates on the obsolete write, which is the
eviction table's failure reproduced at full length. The causal router
also matches the question-holding SnapKV reference on multikey (0.95
for both at 8K) by keeping every needle, whichever key the question
later names.

\paragraph{Shortcut control.} In the register-log training data the
queried register is always the corrected one, so ranking corrections
first is a learnable shortcut. Table~\ref{tab:pq-staleany} repeats the
evaluation with the queried register drawn uniformly over identical
contexts. Below the evidence-set size no query-free policy can cover
every candidate question. At budgets that hold the full set the causal
margin over H2O is 21 points at 8K and 23 at 16K, with no shortcut
available.

% Generated by analysis/make_external_appendix.py from runs/external/. Do not edit by hand.

\setlength{\tabcolsep}{2.6pt}
\begin{table}[H]
\caption{Query-aware eviction (the router and every policy see the question), Qwen2.5-7B-Instruct, 10\% budget. Example counts in parentheses. The evidence-set oracle's 0.93 on the register log is the first-token sink: keeping the first context block as well reaches 1.00.}
\label{tab:qa}
\begin{center}\small
\begin{tabular}{llccccc}
\toprule
Condition & Length & causal & imitation & SnapKV-style & MomentKV-style & evidence set \\
\midrule
single needle & 8K (60) & 1.00 & 0.70 & 1.00 & 0.95 & 1.00 \\
 & 16K (60) & 1.00 & 0.53 & 1.00 & 0.97 & 1.00 \\
\midrule
multikey & 8K (60) & 0.78 & 0.12 & 0.95 & 0.10 & 1.00 \\
 & 16K (60) & 0.78 & 0.13 & 1.00 & 0.15 & 1.00 \\
\midrule
variable tracking & 8K (58) & 0.79 & 0.00 & 0.10 & 0.00 & 0.91 \\
 & 16K (53) & 0.89 & 0.00 & 0.81 & 0.21 & 0.98 \\
\midrule
register log & 8K (60) & 1.00 & 0.07 & 1.00 & 0.17 & 0.93 \\
 & 16K (60) & 1.00 & 0.05 & 1.00 & 0.32 & 1.00 \\
\bottomrule
\end{tabular}
\end{center}
\end{table}

\begin{table}[H]
\caption{Pre-query eviction at 16K (four times the training length), Qwen2.5-7B-Instruct. Fraction of dense-correct examples preserved under each block-retention policy (example counts in parentheses). The two routers differ only in the supervision target. $\dagger$SnapKV reads the question and is a reference from the query-aware regime, not a comparator.}
\label{tab:pq16k}
\begin{center}\small
\begin{tabular}{llcccccc}
\toprule
Condition & Budget & causal & imitation & H2O-style & MomentKV-style & evidence set & SnapKV-style$^\dagger$ \\
\midrule
single needle (60) & 10\% & 1.00 & 0.53 & 0.93 & 0.97 & 1.00 & 1.00 \\
 & 25\% & 1.00 & 0.83 & 1.00 & 1.00 & 1.00 & 1.00 \\
 & 50\% & 1.00 & 0.95 & 1.00 & 1.00 & 1.00 & 1.00 \\
\midrule
multikey (60) & 10\% & 0.97 & 0.15 & 0.15 & 0.15 & 1.00 & 1.00 \\
 & 25\% & 0.98 & 0.28 & 0.30 & 0.35 & 1.00 & 1.00 \\
 & 50\% & 1.00 & 0.60 & 0.58 & 0.53 & 1.00 & 1.00 \\
\midrule
variable tracking (53) & 10\% & 0.85 & 0.00 & 0.04 & 0.21 & 0.79 & 0.81 \\
 & 25\% & 0.92 & 0.04 & 0.72 & 0.60 & 0.79 & 0.79 \\
 & 50\% & 0.91 & 0.34 & 0.85 & 0.85 & 0.79 & 0.92 \\
\midrule
register log (60) & 10\% & 1.00 & 0.05 & 0.32 & 0.30 & 0.32 & 1.00 \\
 & 25\% & 1.00 & 0.28 & 1.00 & 0.45 & 1.00 & 1.00 \\
 & 50\% & 0.98 & 0.45 & 1.00 & 0.67 & 1.00 & 1.00 \\
\bottomrule
\end{tabular}
\end{center}
\end{table}

\begin{table}[H]
\caption{Shortcut control for the register log. Identical contexts, with the queried register drawn uniformly so it is the corrected one on about one example in sixteen. $\dagger$SnapKV reads the question and is a reference, not a comparator.}
\label{tab:pq-staleany}
\begin{center}\small
\begin{tabular}{llcccccc}
\toprule
Length & Budget & causal & imitation & H2O-style & MomentKV-style & evidence set & SnapKV-style$^\dagger$ \\
\midrule
8K (59) & 10\% & 0.34 & 0.20 & 0.25 & 0.20 & 0.44 & 0.92 \\
 & 25\% & 0.75 & 0.49 & 0.54 & 0.54 & 0.78 & 0.97 \\
 & 50\% & 0.98 & 0.69 & 0.76 & 0.80 & 1.00 & 1.00 \\
\midrule
16K (60) & 10\% & 0.65 & 0.25 & 0.25 & 0.33 & 0.62 & 1.00 \\
 & 25\% & 1.00 & 0.40 & 0.77 & 0.75 & 1.00 & 1.00 \\
 & 50\% & 1.00 & 0.78 & 1.00 & 0.83 & 1.00 & 1.00 \\
\midrule
32K (60) & 10\% & 1.00 & 0.25 & 0.37 & 0.33 & 1.00 & 1.00 \\
 & 25\% & 1.00 & 0.48 & 0.88 & 0.83 & 1.00 & 1.00 \\
 & 50\% & 1.00 & 0.65 & 1.00 & 0.97 & 1.00 & 1.00 \\
\bottomrule
\end{tabular}
\end{center}
\end{table}

\paragraph{Seeds and a second scale.} Retraining both routers from two
further initializations moves the causal router by at most five points
in every budget-feasible cell except one. Variable tracking
and the register log at the 10\% budget are budget-infeasible at 8K,
the query-free evidence set exceeding six blocks, so rankings there
select different subsets and vary across seeds. Multikey at the same
budget is feasible and still spreads from 0.95 to 0.78: a genuine seed
effect at the tightest budget, gone by 25\%. Qwen2.5-3B-Instruct reproduces the pattern. Across both scales the
causal router leads H2O in 34 of 48 task-budget cells and ties within
two points in the other 14. Under teacher-forced token match the 3B
results show a single-needle inversion at 16K and the 10\% budget; the
generative rule removes it, so it was a token-match artifact, not an
eviction effect.

\paragraph{Longer context.} The same 7B routers, evaluated at 32K,
hold their 16K values on single retrieval and on both register-log
conditions, and the multikey margin over H2O remains.
At 32K a 10\% budget is about 25 blocks, so
the full register-log evidence set fits for the first time and the
shortcut control becomes feasible there (Table~\ref{tab:pq-staleany},
32K rows). Variable tracking is the exception. The dense model itself
weakens at this length (45 of 60 examples remain generation-correct).
The causal router falls from 0.92 at 16K to 0.67 at the 25\% budget,
where H2O scores 0.71. At the 10\% budget it still scores 0.76 against
0.67 for the truncated evidence set and 0.38 for H2O.

\paragraph{A second family and a larger scale.}
The 7B pattern reproduces on Llama-3.1-8B and on Qwen2.5-32B; the
causal-over-H2O margins sit near 0.8 from 3B through 8B and are largest
at 32B (Figure~\ref{fig:eviction-scale}). At 16K and a 10\%
budget the causal router preserves 0.98 of Qwen-32B's multikey answers
against 0.07 for H2O. The imitation router sits at 0.13. Variable
tracking is 1.00 against 0.00. Llama-8B is the same ordering, 0.90
against 0.10 on multikey. Both models are scored by greedy generation.
Teacher-forced token match undercounts them. Llama phrases the right
value in different tokens, and the 32B model rephrases. Dense
generation is 60 of 60 on every condition. Qwen-7B's weaker multikey
numbers do not reappear; both causal routers hold 0.90 or better. The
numbers above are router seed 0. Evaluating seeds 1 and 2 at 16K moves the
causal router by at most 0.03 on multikey, 0.97--0.98 at 32B and
0.87--0.90 on Llama; variable tracking holds 1.00 on all three 32B
seeds and spans 0.83--0.88 on Llama, with the attention router at its
floor throughout. The shortcut control repeats at 8K for both models and at
16K for Llama (Table~\ref{tab:pq-staleany-fs}).

\paragraph{Physical eviction.} The masked protocol hides dropped blocks
during the context prefill as well, and deployed eviction does not: the
context prefills densely, the discarded entries leave the cache, and the
retained entries were computed with the dropped blocks visible, so
evidence can survive eviction inside them. Table~\ref{tab:evict}
repeats the pre-query evaluation under that protocol, with cached keys
at their original rotary positions and the question appended afterward.
The surviving information helps rather than hurts the causal router: at
16K and the 10\% budget it preserves 0.93 to 1.00 on multikey,
variable tracking, and the register log, and 0.65 on the any-query
control against 0.43 for H2O. Variable tracking rises from 0.85 under
masking to 1.00, since retained neighbors carry fragments of dropped
chain links. H2O gains on some conditions for the same reason and still
trails by 50 points or more on multikey, variable tracking, and the
register log, and the imitation router does not move at all. A masked-prefill variant of this pipeline reproduces the masked
tables exactly, so the two protocols differ only in what the retained
entries saw during prefill.

\paragraph{Output reconstruction as the indicator.} ReST-KV
\citep{an2026restkv} scores a token by the change in the attention
output when its cache entry is removed, with attention through a
nonlinear amplifier as one factor, and evicts once at prefill.
Table~\ref{tab:restkv} transplants that indicator into this harness at
block granularity, with the reconstruction norm computed exactly per
head. Given the question window, it tracks SnapKV within a few points on
retrieval and the register log and leads it on variable tracking at
the 10\% budget, 0.26 against 0.10, so the transplant preserves the
indicator's strength. On context rows alone it falls to the level of accumulated
attention: 0.07 against H2O's 0.11 on multikey at 8K and the 10\%
budget, and 0.00 on variable tracking. A reconstruction signal computed
before the question exists measures how the context predicts itself,
not what a future question will need. Under this matched global-block
protocol the tested passive-context readouts, accumulated attention,
window attention, and output reconstruction, are poor pre-query
targets. Scores from a different scoring task escape the pattern:
KVzip's repeat-pass score, measured below, preserves \KvzipTwoFive{}
of QASPER answers at the 25\% budget (Table~\ref{tab:kvzip}). Head-
and layer-specific learned policies
\citep{jegou2026kvzap, moschella2026kvpolicy} remain untested here.

% Generated by analysis/make_external_appendix.py from runs/external/. Do not edit by hand.

\begin{table}[H]
\caption{Shortcut control at the second family and the larger scale, scored by greedy generation. Same construction as Table~\ref{tab:pq-staleany}. $\dagger$SnapKV reads the question and is a reference, not a comparator.}
\label{tab:pq-staleany-fs}
\begin{center}\small
\begin{tabular}{llcccccc}
\toprule
Model, length & Budget & causal & imitation & H2O-style & MomentKV-style & evidence set & SnapKV-style$^\dagger$ \\
\midrule
Llama-8B, 8K (60) & 10\% & 0.48 & 0.30 & 0.30 & 0.27 & 0.43 & 0.98 \\
 & 25\% & 0.90 & 0.42 & 0.53 & 0.47 & 0.85 & 1.00 \\
 & 50\% & 1.00 & 0.75 & 0.77 & 0.80 & 1.00 & 1.00 \\
\midrule
Llama-8B, 16K (60) & 10\% & 0.68 & 0.25 & 0.28 & 0.37 & 0.60 & 0.98 \\
 & 25\% & 1.00 & 0.48 & 0.75 & 0.65 & 0.97 & 0.98 \\
 & 50\% & 1.00 & 0.65 & 0.98 & 0.93 & 1.00 & 0.98 \\
\midrule
Qwen-32B, 8K (60) & 10\% & 0.35 & 0.20 & 0.25 & 0.15 & 0.43 & 0.83 \\
 & 25\% & 0.77 & 0.43 & 0.50 & 0.38 & 0.78 & 0.98 \\
 & 50\% & 1.00 & 0.70 & 0.70 & 0.83 & 1.00 & 1.00 \\
\bottomrule
\end{tabular}
\end{center}
\end{table}

\begin{table}[H]
\caption{The output-reconstruction indicator of \citet{an2026restkv} transplanted to block eviction, Qwen2.5-7B-Instruct. Divergences from the cited method: blocks instead of tokens, one global selection instead of per-head and per-layer, a window mean in place of the smoothing, and context-row queries in the pre-query columns. The indicator itself is unmodified.}
\label{tab:restkv}
\begin{center}\small
\begin{tabular}{llcccccc}
\toprule
 & & \multicolumn{3}{c}{pre-query} & \multicolumn{3}{c}{query-aware} \\
Condition & Length & 10\% & 25\% & 50\% & 10\% & 25\% & 50\% \\
\midrule
single needle & 8K (60) & 0.83 & 0.92 & 0.97 & 1.00 & 1.00 & 1.00 \\
 & 16K (60) & 0.92 & 0.93 & 0.97 & 1.00 & 1.00 & 1.00 \\
\midrule
multikey & 8K (60) & 0.07 & 0.20 & 0.42 & 0.95 & 1.00 & 1.00 \\
 & 16K (60) & 0.13 & 0.25 & 0.50 & 1.00 & 1.00 & 1.00 \\
\midrule
variable tracking & 8K (58) & 0.00 & 0.07 & 0.24 & 0.26 & 0.97 & 0.97 \\
 & 16K (53) & 0.09 & 0.34 & 0.60 & 0.81 & 0.79 & 0.92 \\
\midrule
register log & 8K (60) & 0.15 & 0.38 & 0.62 & 1.00 & 1.00 & 1.00 \\
 & 16K (60) & 0.60 & 0.67 & 0.82 & 1.00 & 1.00 & 1.00 \\
\midrule
register log, any query & 8K (59) & 0.24 & 0.39 & 0.75 & 0.90 & 0.95 & 1.00 \\
 & 16K (59) & 0.34 & 0.61 & 0.73 & 0.97 & 1.00 & 1.00 \\
\bottomrule
\end{tabular}
\end{center}
\end{table}

\begin{table}[H]
\caption{Physical KV eviction, Qwen2.5-7B-Instruct at the 10\% budget: the context prefills densely, the cache entries of unselected blocks are removed before the question exists, and the question then attends to the pruned cache at original positions. Same data, routers, and scoring as the masked tables, on the examples dense-correct under both protocols.}
\label{tab:evict}
\begin{center}\small
\begin{tabular}{llcccccc}
\toprule
Condition & Length & causal & imitation & H2O-style & recency & random & evidence set \\
\midrule
single needle & 8K (60) & 1.00 & 0.68 & 0.87 & 0.13 & 0.12 & 1.00 \\
 & 16K (60) & 1.00 & 0.53 & 0.95 & 0.12 & 0.08 & 1.00 \\
\midrule
multikey & 8K (60) & 0.97 & 0.12 & 0.10 & 0.13 & 0.18 & 1.00 \\
 & 16K (60) & 0.93 & 0.15 & 0.13 & 0.07 & 0.08 & 0.98 \\
\midrule
variable tracking & 8K (58) & 0.47 & 0.00 & 0.00 & 0.00 & 0.00 & 0.00 \\
 & 16K (53) & 1.00 & 0.00 & 0.30 & 0.00 & 0.00 & 0.98 \\
\midrule
register log & 8K (60) & 0.95 & 0.10 & 0.20 & 0.43 & 0.13 & 0.13 \\
 & 16K (60) & 1.00 & 0.05 & 0.42 & 0.42 & 0.05 & 0.32 \\
\midrule
register log, any query & 8K (59) & 0.36 & 0.20 & 0.32 & 0.15 & 0.08 & 0.44 \\
 & 16K (60) & 0.65 & 0.25 & 0.43 & 0.12 & 0.10 & 0.62 \\
\bottomrule
\end{tabular}
\end{center}
\end{table}

\paragraph{Natural text.} QASPER \citep{dasigi2021qasper} provides
research papers with reader-written questions and annotator-marked
evidence paragraphs. Surface tokens do not bind a question to its
evidence. We treat paragraphs as blocks and give every policy a token
budget rather than a block count. Questions are the extractive and
yes/no ones whose evidence is full-text paragraphs: \QasperRows{}
questions over \QasperDocs{} held-out documents, of which the dense
model answers \QasperDense{} correctly, over \QasperCondDocs{}
documents, and every table conditions on those. An answer counts as contained when its normalized word sequence
appears contiguously in a short greedy continuation decoded with no
access to the reference; the continuations ship in the results files.
Pre-query training targets are the union of a paper's per-question
labels, and one selection per document serves all of its questions; the
shipped results verify the kept sets are identical across a document's
questions at every budget.
The annotated evidence union for a paper fits a 25\% token budget and
preserves \QasperRefB{} of answers before any question is known.
The reference measures selection that
prioritizes annotated evidence; it is not an upper bound on the frozen
model under selection, and the register-log analogue rises from 0.93
to 1.00 when the first context block is kept as well
(Table~\ref{tab:qa}). H2O preserves \QasperHtwoB{} at the
same budget, below the \QasperRandB{} of random paragraphs, so
accumulated attention points away from evidence on natural text as it
does on the register log. A matched router pair trained on intervention
labels from the training split reaches \QasperBaseRcB{} at the 25\%
budget, ahead of every attention-derived policy.
At the 10\% budget its recall of the
annotated evidence on unseen papers is 0.16 against 0.87 on training
papers, so the remaining gap to the annotated-evidence reference is a
generalization problem in the selector. SnapKV reaches \QasperSnapB{}
once it can read the question.

\paragraph{A stronger selector.} Replacing the pooled-embedding
features with frozen Qwen2.5-0.5B paragraph encodings, and keeping the
mixer, training, and document split fixed, lifts the matched pair
(Table~\ref{tab:qasper-pair}, router seed 0; across three seeds
recovered spans \QasperPairRcSeedLo{} to \QasperPairRcSeedHi{} and
attention \QasperPairRaSeedLo{} to \QasperPairRaSeedHi{} at the 25\%
budget). Recovered intervention labels reach \QasperPairRcB{} there
against \QasperPairRaB{} for attention labels; the paired difference,
clustered by document, is
$+0.18$ [$+0.10$, $+0.27$] and excludes zero at every budget, against
$+0.10$ for the pooled-feature pair. Recovered labels sit eleven
points under the annotated diagnostic at the 10\% budget, five points
under at 25\%, and above it at 50\%, so recovering the target costs
most where the budget is tightest. The distance that remains to the
\QasperRefB{} annotated-evidence reference belongs to query-free
selection on unseen documents, with the annotated arm at
\QasperPairAnnB{} at the 25\% budget and \QasperPairAnnBfifty{} at
50\%. Under physical KV eviction the frozen selectors
reproduce their masked results within three points at both feasible
budgets. An alternative answer-aware scorer,
first-token parsing for yes and no answers and token F1 of at least
0.5 for extractive spans, is not uniformly stricter than containment;
on the same \QasperDense{} dense-correct questions it leaves the
ordering unchanged at the 25\% budget, \QasperStrictRc{} recovered
against \QasperStrictRa{} attention.

\paragraph{All eligible questions.} The tables above condition on
dense-correct questions. Table~\ref{tab:qasper-all} drops that filter
and scores every eligible question under physical eviction. Yes/no
answers are read from the first yes or no token
(\AllYesNoN{} questions). Extractive answers get token F1 against each
annotator's spans, keeping the best match (\AllExtractiveN{} questions).
On yes/no questions the strict parse and containment
agree on every continuation in every arm, and no continuation contains
both words. At the 25\% budget the recovered-label selector leads the
attention-label selector by \AllRaFOneTwoFive{} in extractive F1 and
\AllRaContainTwoFive{} in containment, clustered by document, and by
\AllRaFOneFifty{} in F1 at 50\%. On the yes/no questions the two do not
separate (\AllRaYesNoTwoFive{}). KVzip leads the selector on every
metric at both budgets. The selector-minus-KVzip F1 difference is
\AllRkFOneTwoFive{} at 25\% and \AllRkFOneFifty{} at 50\%. The dense model often answers a yes/no question in prose without
either word, which is why its strict yes/no score is low.

\paragraph{Selector training protocol.} The label pass runs on the
training split: \QasperTrainQs{} eligible questions over
\QasperTrainDocs{} documents, of which the scoring model answers
\QasperTrainSolved{} correctly under teacher forcing; the estimator
returns a recovered set on \QasperTrainLabeledQs{} of those, and their
per-question sets union into document-level targets on
\QasperTrainLabeledDocs{} documents. Each question contributes one
reference answer, the first annotator answer whose evidence is
full-text paragraphs, with extractive answers taking the first marked
span and yes/no answers rendered as words; further reference answers
are unused. The intervention score is the mean log-probability of the
reference answer's tokens under teacher forcing, so multi-token
answers average over their tokens. The frozen encoder is
Qwen2.5-0.5B-Instruct: each paragraph is encoded independently,
truncated at 512 tokens, its layer-12 hidden states mean-pooled, and a
normalized paragraph-position scalar appended; a two-layer transformer
mixer over the paragraph sequence scores each paragraph, and
paragraphs are kept greedily under the token budget.

\paragraph{External compression.} LLMLingua-2 \citep{pan2024llmlingua2}
deletes tokens with a distilled classifier and needs no question, so it
enters the pre-query regime as an external system rather than a matched
arm. On the same dense-correct questions it reaches \QasperLLTen{},
\QasperLLTwentyFive{}, and \QasperLLFifty{} at the three budgets
(Table~\ref{tab:qasper-pair}). Against the recovered-label selector
the paired differences, clustered by document, are $-0.03$
[$-0.11, +0.04$] at 10\%, $+0.04$ [$-0.04, +0.12$] at 25\%, and
$+0.10$ [$+0.03, +0.18$] at 50\%: indistinguishable at the two
tighter budgets and a selector lead at half retention. Token
deletion can keep fragments from every paragraph where paragraph
retention commits to whole blocks, which may explain the tight-budget
pattern; these runs do not test that mechanism. Both approaches leave
the annotated-evidence reference well above them.

\paragraph{The official KVzip system.} KVzip \citep{kim2025kvzip}
scores a cached KV pair by the maximum attention it receives while the
model repeats the context back, and evicts per head at pair
granularity. We run the released implementation unmodified as the
compressor, on the same checkpoint, prompts, decode lengths, and
containment scorer as the shipped tables, under physical eviction: the
title header is the protected prefix, the question never reaches the
compressor, and every question of a document starts from the same
compressed cache. On the \KvzipCohortN{} questions over
\KvzipCohortDocs{} documents that the dense model answers correctly
under this protocol, KVzip preserves \KvzipTwoFive{} at the 25\%
budget and \KvzipFifty{} at 50\%, against \KvzipRecTwoFive{} and
\KvzipRecFifty{} for the recovered-label selector; the paired
differences, clustered by document, are \KvzipPairTwoFive{} at 25\% and
\KvzipPairFifty{} at 50\% (Table~\ref{tab:kvzip}). The ordering survives
the alternative answer-aware scorer and holds on yes/no and extractive
questions separately. On the 16K multikey and variable-tracking
conditions (\KvzipMkN{} and \KvzipVtN{} dense-correct examples), KVzip
preserves \KvzipMkTen{} and \KvzipVtTen{} at the 10\% budget where
the causal router preserves \RouterMkTen{} and \RouterVtTen{}, and
\KvzipMkTwoFive{}/\KvzipVtTwoFive{} against
\RouterMkTwoFive{}/\RouterVtTwoFive{} at 25\%: similar point
estimates with no statistically resolved difference. The comparison
changes the scoring task and the selection granularity together, and a
trained selector must generalize to unseen documents where KVzip scores
each document directly.

\paragraph{KVzip on the register log.} Native KVzip preserves the current
value on every dense-correct example on the register log and the shortcut
control, at 8K and 16K and both the 10\% and 25\% budgets
(Table~\ref{tab:kvzip-stale}). On the register log the causal router
preserves \KzStaleCausalLo{} to \KzStaleCausalHi{} and H2O-style eviction
\KzStaleHtwoLo{} to \KzStaleHtwoHi{}. Evidence labels lead
attention-based eviction on that task. They do not lead reconstruction
scoring.

\paragraph{Selection granularity.}
Native KVzip evicts per head at pair granularity. Reduced to
paragraphs, each paragraph's score is the fraction of its KV pairs
that KVzip keeps at the budget, under KVzip's own global threshold
(Tables~\ref{tab:qasper-kvzip-labels}
and~\ref{tab:kvzip-granularity}). Ranking whole paragraphs by that
score, with no learned selector, preserves \GranDirectTwoFive{} at
25\% against native KVzip's \KvzipTwoFive{}. The paired difference is
\GranDirectNativeTwoFive{}, clustered by document. The same selector
trained on those paragraph labels, sized to match each recovered
label, trails the recovered-label selector by \MaskedSetRecHi{} to
\MaskedSetRecLo{} at 25\% in the masked protocol across router seeds,
and by \GranSetRecTwoFive{} under physical eviction. It matches
ranking the paragraphs directly. A soft target built from KVzip
scores does no better than attention labels. Intervention labels are
the strongest paragraph-level target we test.

\paragraph{Sentence units.} Finer units help only modestly. With each
paragraph split into sentences and the prompt text unchanged, the same
arms run at both granularities on the same \SentN{} dense-correct
questions under the masked protocol (Table~\ref{tab:qasper-sentences}),
with kept tokens matched within \SentKeptPct{}\%. The annotated-evidence
oracle shows no detectable improvement at 25\% or 50\% and declines at
10\%. The KVzip ranking gains \SentKzTwoFive{} at 25\% and
\SentKzFifty{} at 50\%, and the selector trained on annotated labels gains
\SentAnnTwoFive{} and \SentAnnFifty{}, clustered by document.
Sentence-level recovered-label and attention-label selectors were not
trained. In a pilot of 100 training papers, \SentPilotOutside{}\% of
recovered sentences fell outside annotated evidence paragraphs. Under physical eviction native KVzip keeps its lead
(Table~\ref{tab:qasper-sentences-evict}). Sentence-level ranking of
KVzip's scores does not close it. The annotated-evidence oracle at
sentence units matches native KVzip at 25\%.

Validation: the achieved retained fraction of context KV pairs matches
the nominal budget for both methods (0.250 and 0.500 for KVzip; 0.247
and 0.498 for paragraph selection on the audited subset); KVzip's
pipeline agrees with the shipped harness on dense-generation
correctness for \KvzipDenseAgree{} of \KvzipDenseTotal{} questions,
and restricting to the \CrossBothN{} questions dense-correct under both
paths leaves every ordering unchanged; pruning at ratio 1.0 leaves
generations byte-identical on \KvzipIdentSame{} of \KvzipIdentTotal{}
audited documents, the remainder differing by one greedy token from the
eviction kernel path; and pushing identical paragraph keep-sets through
both decode backends agrees on correctness for \CrossAgree{} of
\CrossRows{} question-budget pairs with a net difference of
\CrossNetBias{}, an order of magnitude below the system difference.
Scoring costs a median \KvzipScoreSecs{} seconds per document on one
RTX 6000 Pro, amortized over the document's future questions.

% Generated by analysis/make_external_appendix.py from runs/external/. Do not edit by hand.

\setlength{\tabcolsep}{2.6pt}
\begin{table}[H]
\caption{The same selector class with frozen-encoder paragraph features, trained under three label sources with everything else fixed, on the dense-correct questions of the held-out documents. The annotated arm is a learnability diagnostic; recovered labels come from the intervention estimator. $\ddagger$LLMLingua-2 is an external system that edits the text at token level with the question withheld, not a matched arm; its achieved token ratios are 0.10, 0.25, and 0.50.}
\label{tab:qasper-pair}
\begin{center}\small
\begin{tabular}{lcccc}
\toprule
Budget & annotated & recovered & attention & LLMLingua-2$^\ddagger$ \\
\midrule
10\% & 0.43 & 0.32 & 0.19 & 0.35 \\
25\% & 0.59 & 0.55 & 0.36 & 0.50 \\
50\% & 0.71 & 0.73 & 0.54 & 0.63 \\
\bottomrule
\end{tabular}
\end{center}
\end{table}

\begin{table}[H]
\caption{The official KVzip system against the frozen selectors under physical KV eviction on QASPER. Cohort: the 238 questions over 164 documents the dense model answers correctly under the eviction protocol, router seed 0. KVzip scores with its native chunked repeat pass and evicts per head at pair granularity; the selectors commit to whole paragraphs. The last column is the paired difference over questions with a 95\% bootstrap interval clustered by document.}
\label{tab:kvzip}
\begin{center}\small
\begin{tabular}{lcccc}
\toprule
Budget & KVzip & recovered & attention & KVzip $-$ recovered \\
\midrule
25\% & 0.86 & 0.51 & 0.34 & $+0.34$ [$+0.27$, $+0.42$] \\
50\% & 0.92 & 0.71 & 0.55 & $+0.21$ [$+0.15$, $+0.28$] \\
\bottomrule
\end{tabular}
\end{center}
\end{table}

\begin{table}[H]
\caption{The selector of Table~\ref{tab:qasper-pair} trained on targets taken from KVzip's reconstruction scores, masked protocol, on the same 234 dense-correct questions. KVzip set keeps, per training paper, as many top paragraphs by KVzip retention as the recovered label has; KVzip soft distills each paragraph's share of total KVzip score, the form of the attention target. KVzip ranking uses no selector: it keeps the dev paragraphs KVzip itself retains most of. Means over three router seeds (KVzip ranking has none); seeds differ by at most 0.04.}
\label{tab:qasper-kvzip-labels}
\begin{center}\small
\begin{tabular}{lccccc}
\toprule
Budget & recovered & attention & KVzip set & KVzip soft & KVzip ranking \\
\midrule
10\% & 0.33 & 0.19 & 0.27 & 0.25 & 0.26 \\
25\% & 0.54 & 0.37 & 0.43 & 0.38 & 0.47 \\
50\% & 0.73 & 0.55 & 0.66 & 0.57 & 0.67 \\
\bottomrule
\end{tabular}
\end{center}
\end{table}

\begin{table}[H]
\caption{Where the KVzip advantage comes from, under physical eviction on the 238 questions of Table~\ref{tab:kvzip}, router seed 0. KVzip pairs is the native system, evicting per head at pair granularity. KVzip paragraphs keeps whole paragraphs ranked by the fraction of their pairs KVzip retains. The two selectors are trained on KVzip-derived and on recovered paragraph labels.}
\label{tab:kvzip-granularity}
\begin{center}\small
\begin{tabular}{lcccc}
\toprule
 & \multicolumn{2}{c}{KVzip scores} & \multicolumn{2}{c}{selector labels} \\
Budget & pairs & paragraphs & KVzip & recovered \\
\midrule
25\% & 0.86 & 0.49 & 0.44 & 0.51 \\
50\% & 0.92 & 0.66 & 0.66 & 0.71 \\
\bottomrule
\end{tabular}
\end{center}
\end{table}

\begin{table}[H]
\caption{Sentence against paragraph units under the masked protocol, on the 234 dense-correct questions over 161 dev documents. Each paragraph is split into sentences with the prompt text unchanged, so both granularities run on the same questions and keep the same token budget. The last column is the paired difference with a 95\% bootstrap interval clustered by document.}
\label{tab:qasper-sentences}
\begin{center}\small
\begin{tabular}{llccc}
\toprule
Arm & Budget & paragraph & sentence & sentence $-$ paragraph \\
\midrule
annotated-evidence oracle & 10\% & 0.86 & 0.82 & $-0.04$ [$-0.07$, $-0.01$] \\
 & 25\% & 0.84 & 0.85 & $+0.01$ [$-0.02$, $+0.05$] \\
 & 50\% & 0.86 & 0.87 & $+0.00$ [$-0.03$, $+0.03$] \\
\midrule
annotated-label selector & 10\% & 0.43 & 0.41 & $-0.01$ [$-0.09$, $+0.07$] \\
 & 25\% & 0.59 & 0.65 & $+0.06$ [$-0.02$, $+0.13$] \\
 & 50\% & 0.71 & 0.79 & $+0.08$ [$+0.02$, $+0.14$] \\
\midrule
KVzip ranking & 10\% & 0.26 & 0.28 & $+0.02$ [$-0.03$, $+0.07$] \\
 & 25\% & 0.47 & 0.54 & $+0.07$ [$+0.00$, $+0.13$] \\
 & 50\% & 0.67 & 0.74 & $+0.07$ [$+0.02$, $+0.12$] \\
\bottomrule
\end{tabular}
\end{center}
\end{table}

\begin{table}[H]
\caption{Native KVzip on the register log and its shortcut control, beside the router and eviction arms of Table~\ref{tab:evict} on the same examples (dense prefill, physical eviction; examples the dense model answers correctly under both pipelines). Cells are the share whose answer contains the current value.}
\label{tab:kvzip-stale}
\begin{center}\small
\begin{tabular}{lllcccc}
\toprule
Condition & Length & Budget & KVzip & causal & imitation & H2O-style \\
\midrule
register log & 8K (60) & 10\% & 1.00 & 0.95 & 0.10 & 0.20 \\
 &  & 25\% & 1.00 & 1.00 & 0.20 & 0.33 \\
 & 16K (60) & 10\% & 1.00 & 1.00 & 0.05 & 0.42 \\
 &  & 25\% & 1.00 & 1.00 & 0.28 & 0.57 \\
\midrule
shortcut control & 8K (59) & 10\% & 1.00 & 0.36 & 0.20 & 0.32 \\
 &  & 25\% & 1.00 & 0.75 & 0.49 & 0.51 \\
 & 16K (60) & 10\% & 1.00 & 0.65 & 0.25 & 0.43 \\
 &  & 25\% & 1.00 & 1.00 & 0.40 & 0.78 \\
\bottomrule
\end{tabular}
\end{center}
\end{table}

\begin{table}[H]
\caption{The keep-sets of Table~\ref{tab:qasper-sentences} replayed under physical eviction on the 238 questions of Table~\ref{tab:kvzip}, beside native KVzip. The last column is the sentence arm minus native KVzip, with a 95\% bootstrap interval clustered by document.}
\label{tab:qasper-sentences-evict}
\begin{center}\small
\begin{tabular}{llcccc}
\toprule
Arm & Budget & paragraph & sentence & native KVzip & sentence $-$ KVzip \\
\midrule
annotated-evidence oracle & 25\% & 0.85 & 0.88 & 0.86 & $+0.02$ [$-0.03$, $+0.07$] \\
 & 50\% & 0.87 & 0.87 & 0.92 & $-0.05$ [$-0.10$, $-0.00$] \\
\midrule
annotated-label selector & 25\% & 0.56 & 0.63 & 0.86 & $-0.22$ [$-0.29$, $-0.16$] \\
 & 50\% & 0.73 & 0.81 & 0.92 & $-0.11$ [$-0.17$, $-0.05$] \\
\midrule
KVzip ranking & 25\% & 0.49 & 0.53 & 0.86 & $-0.32$ [$-0.40$, $-0.25$] \\
 & 50\% & 0.66 & 0.73 & 0.92 & $-0.20$ [$-0.26$, $-0.14$] \\
\bottomrule
\end{tabular}
\end{center}
\end{table}

\begin{table}[H]
\caption{QASPER under physical eviction, scored on all 658 eligible questions with no dense-correct filter. Strict yes/no reads the first yes or no token (89 questions); extractive F1 is the best token F1 over annotators (569 questions).}
\label{tab:qasper-all}
\begin{center}\small
\begin{tabular}{llccc}
\toprule
Budget & Arm & Containment & Strict yes/no & Extractive F1 \\
\midrule
-- & dense & 0.36 & 0.38 & 0.30 \\
\midrule
25\% & recovered & 0.23 & 0.24 & 0.22 \\
25\% & attention & 0.15 & 0.25 & 0.16 \\
25\% & KVzip & 0.35 & 0.51 & 0.32 \\
\midrule
50\% & recovered & 0.28 & 0.28 & 0.27 \\
50\% & attention & 0.24 & 0.33 & 0.23 \\
50\% & KVzip & 0.36 & 0.40 & 0.32 \\
\bottomrule
\end{tabular}
\end{center}
\end{table}

\paragraph{Scope.} The RULER conditions are lexically transparent:
needles share tokens with their questions, which favors both the window
statistic and the router's features. QASPER removes that transparency
for the diagnosis and the annotated-evidence reference. The learned selector there reaches 0.55 against the 0.83 annotated-evidence reference at the 25\% budget. The paper's pretrained multi-hop task remains
unsolved by every router here. Variable tracking has a lexical chain
that task deliberately lacks; its 32K decline is where chain evidence
outgrows these routers.

\end{document}